\documentclass[10pt,twocolumn,letterpaper]{article}
\usepackage[pagenumbers]{cvpr}

\usepackage{multirow}
\usepackage{enumitem}
\usepackage[normalem]{ulem}
\usepackage[title]{appendix}
\usepackage[dvipsnames]{xcolor}

\newcommand{\norm}[1]{\left\lVert#1\right\rVert}
\definecolor{bronze}{rgb}{1,1,0.6}
\definecolor{silve}{rgb}{0.969,0.796,0.600}
\definecolor{gold}{rgb}{0.941,0.592,0.600}
\newcommand{\gold}[1]{\textbf{#1}}

\definecolor{cvprblue}{rgb}{0.21,0.49,0.74}
\usepackage[pagebackref,breaklinks,colorlinks,citecolor=cvprblue]{hyperref}

\def\confName{3DV\xspace}
\def\confYear{2025\xspace}

\title{\hspace{-7pt}Betsu-Betsu: Multi-View Separable 3D Reconstruction of Two Interacting Objects}

\author{Suhas Gopal\textsuperscript{1} \quad Rishabh Dabral\textsuperscript{2} \quad Vladislav Golyanik\textsuperscript{2} \quad Christian Theobalt\textsuperscript{2}\\
\vspace{5pt}
\textsuperscript{1}Saarland University, SIC \quad \textsuperscript{2}Max Planck Institute for Informatics, SIC}

\begin{document}
\twocolumn[{ 
\renewcommand\twocolumn[1][]{#1} 
\maketitle
\begin{center}
  \vspace{-1em}
  \includegraphics[width=\textwidth]{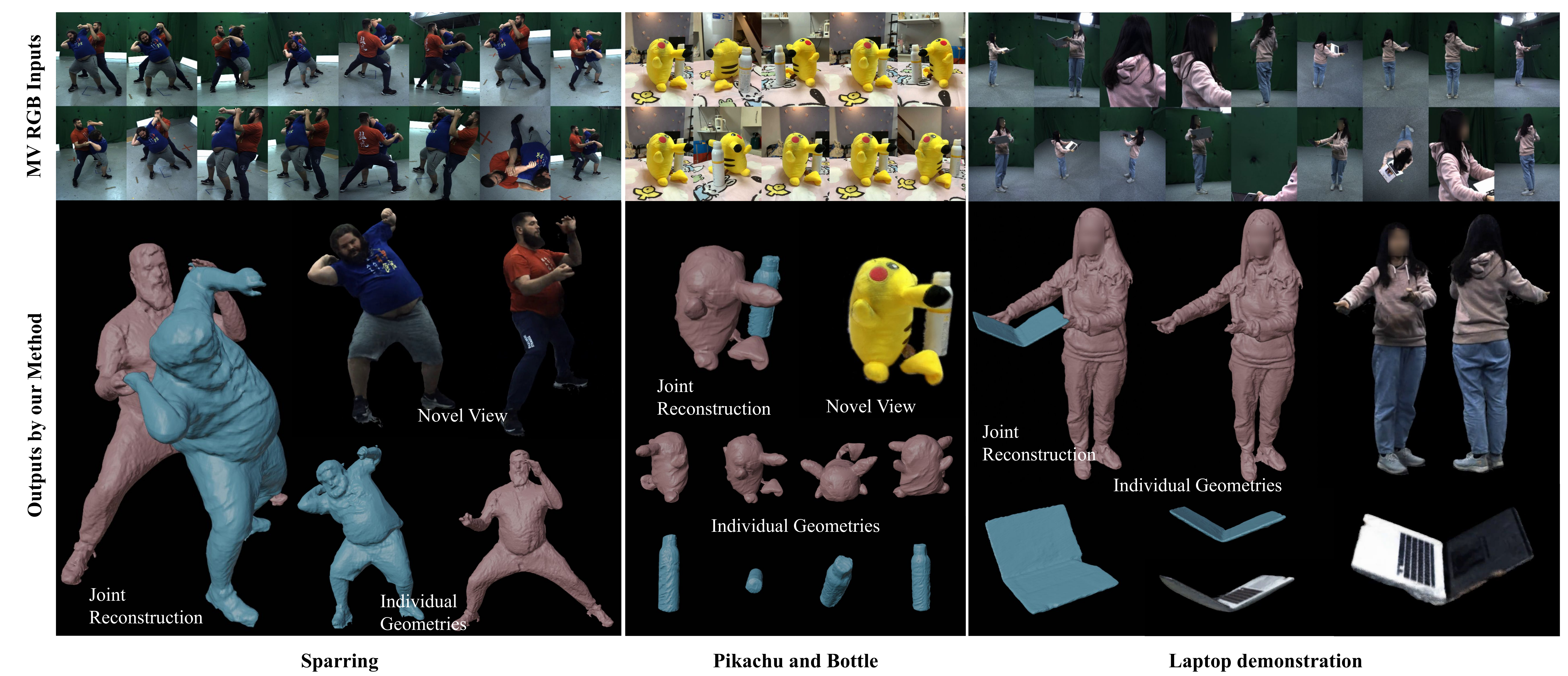}
  \captionof{figure}{
  \textbf{Our method reconstructs humans and objects in 3D from segmented multi-view (MV) RGB images (top) in a separable way}, i.e.~with clean boundaries and no inter-penetration. 
   (Bottom:) For each of the three scenes (\textit{Sparring}, \textit{Pikachu} and \textit{Laptop Demonstration}), we show the two join recovered geometries 
   (left), individual novel view renderings (top right) and individual geometries. 
    }
  \label{fig:teaser}
\end{center}
}] 
\begin{abstract}
Separable 3D reconstruction of multiple objects from multi-view RGB images---resulting in two different 3D shapes for the two objects with a clear separation between them---remains a sparsely researched problem. 
It is challenging due to severe mutual occlusions and ambiguities along the objects' interaction boundaries. 
This paper investigates the setting and 
introduces a new neuro-implicit method that can reconstruct the geometry and appearance of two objects undergoing close interactions 
while disjoining 
both in 3D, avoiding surface inter-penetrations and enabling novel-view synthesis of the observed scene. 
The framework is end-to-end trainable and supervised using a novel alpha-blending regularisation that ensures that the two geometries are well separated even under extreme occlusions. 
Our reconstruction method is markerless and can be applied to rigid as well as articulated objects.
We introduce a new dataset consisting of close interactions between a human and an object and also evaluate on two scenes of humans performing martial arts. 
The experiments confirm the effectiveness of our framework and substantial improvements using 3D and novel view synthesis metrics compared to several existing approaches applicable in our setting\footnote{Project page: \url{https://vcai.mpi-inf.mpg.de/projects/separable-recon/}}. 
\end{abstract}

\section{Introduction}\label{sec:intro}

The world we live in is compositional.
A typical office desk, for example, would consist of a monitor, a keyboard, a few cups of coffee, mobile phones, and so on.
Needless to say, we rarely encounter scenes comprising of one and only one object.
Yet, most 3D reconstruction research~\cite{mildenhall2020nerf, zhang2021stnerf, weng2022humannerf, rosu2023permutosdf, yariv2021volume, wang2021neus} has focused on scenes with only one object (e.g.~the famous caterpillar scene).
When more than one object is present in the scene (e.g.~the GTA Truck scene), the compositionality of the scene is ignored and the entire scene is reconstructed jointly.
\par
Recent works that addressed the challenge of compositional scene reconstruction have either used object templates~\cite{zhang2023neuraldome, bhatnagar22behave, fan2023arctic, GRAB:2020}, or parametric models of humans or hands~\cite{huang2022hhor, ye2022ihoi, Hi4D}.
A few works that propose a \textit{generalised} solution~\cite{wu2022object, Wu2023objectsdfplus} suffer from inter-penetration of the two or more interacting geometries.
In this work, we focus on the generalised compositional reconstruction setting (as shown in~\cref{fig:teaser}) while mitigating the penetration artefacts of the existing literature.
This mandates addressing the challenges posed by severe occlusion of the objects during interaction (e.g.~a person holding a cup), as well as accounting for the difference in object scales while sampling.

\par
With these considerations in perspective, we propose a new markerless, template-free approach for compositional 3D reconstruction of arbitrary objects undergoing interactions in a scene observed from 
multiple views. 
We represent the object geometries as separate Signed Distance Fields (SDFs) and the appearance with the corresponding Neural Radiance Fields. 
The 3D scene is encoded jointly for the objects by using a \textit{shared} multi-resolution hashgrid~\cite{mueller2022instant} which can be decoded into separate SDFs of the  target objects (e.g.~a hand and a book). 
Crucially, we propose a novel alpha-blending loss which enforces that a point lying inside one object has a high opacity only for the corresponding object's SDF while suppressing the opacity for the other. 
This incentivises clean separation boundaries and reduces the penetration volume between the two SDFs, even if the queried point is poorly observed (e.g.~due to occlusion). 
\par
To demonstrate the effectiveness of our method, we capture a new real-world dataset consisting of several scenes of human-object
interactions. 
For this, we ask the subjects to naturally interact with various small and mid-sized objects in a large capture dome. 
In summary, the technical contributions of this paper are as follows: 
\begin{itemize}[noitemsep]
    \item A novel markerless and category-agnostic approach for high-quality 3D reconstruction of two interacting objects from multi-view RGB inputs; 
    \item A shared neuro-implicit representation that can be jointly optimised for the geometries of interacting objects while also supporting separable free-viewpoint rendering; 
    \item An interaction-aware alpha-compositing of opacity values for each SDF enforcing clean separation boundaries and mitigating inter-object 
    penetration in 3D; 
    \item A new multi-view dataset for human-object interactions. 
\end{itemize} 

In addition to the captured dataset, we also evaluate our method on the publicly available WildRGB-D~\cite{xia2024rgbd} (object-object) and AffordPose~\cite{affordpose} (hand-object) datasets, and 
demonstrate its effectiveness on marker-based 3D reconstruction datasets like NeuralDome~\cite{zhang2023neuraldome}. 
We also evaluate human-human interactions on scenes of the ReMoCap~\cite{ghosh2024remos} dataset. 
These scenes involve practitioners performing martial arts poses, thereby leading to challenging interactions. 
The proposed approach performs better than previous state-of-the-art methods such as ObjectSDF++~\cite{Wu2023objectsdfplus} and NeuS2~\cite{neus2}. 
Although not the main objective of this work, we also observe better performance on the related task of segmented novel-view synthesis \cite{mildenhall2020nerf, mueller2022instant}. 

\section{Related Works}
\label{sec:related_works}

We discuss the related works from three perspectives: 
(1) neural scene representations, (3) implicit models for multi-object segmentation and reconstruction and, finally, (3) generic human-object, hand-object and human-human interaction works. 

\subsection{Neural Scene Representations}

Recent advances in neural implicit representations and NeRF-based techniques \cite{mildenhall2020nerf, Tewari2022NeuRendSTAR} have enabled high-quality novel view synthesis and reconstruction of complex scenes from multi-view images. 
Extensions of those for surface reconstruction by \cite{wang2021neus,yariv2021volume,Oechsle2021ICCV_unisurf} and enhancements in speed by works like \cite{mueller2022instant, neus2, rosu2023permutosdf} have shown that it is possible to reconstruct high-quality geometry, in a reasonable amount of time, such that it can be applied to even short videos on per-frame basis.
They have also been 
applied for human rendering \cite{peng2021neuralbody,liu2021neuralactor,weng2022humannerf,zhao2022instantnsr, sun2024metacap}, that extend to dynamic scenes as well as provide pose-conditioned animation capabilities. 
Some works also extend it to multi-person scenarios~\cite{shuai2022multinb,zhang2021stnerf,Menapace2024ToG}.
The most relevant work to ours  \cite{zhang2023neuraldome, sun2021hoifvv} also uses implicit representation for the reconstruction of human-object interaction. 
They both use an SMPL prior \cite{SMPL:2015, SMPL-X:2019} for the human body and an object template for a layer-wise representation. 
HOI-FVV~\cite{sun2021hoifvv} uses sparse-view RGB input to predict occupancy values of the human undergoing interaction with objects.
However, the sparse inputs limit the reconstruction quality, and the object geometries have to be tracked assuming a template is available. 
On the other hand, Zhang et al.~\cite{zhang2023neuraldome} use dense RGB inputs, and obtain separate NeRFs for both human and object, using SMPL and an object template as priors, which are then blended 
to obtain the final reconstruction. 
In contrast, we implicitly learn to separate the two objects by using only the image segmentation masks as additional supervision. 
\par
While similar to Neus2~\cite{neus2} in employing hashgrid encoding with NeuS, we do not adopt their approximate second-derivative formulation. Instead of the coarse-to-fine training strategy, which caused missing reconstructions for small objects, we optimise all hashgrid levels from the start. Additionally, we condition the separate SDF MLP heads on hashgrid feature vectors derived via another MLP.

\subsection{Multi-Object Segmentation/Reconstruction}
While most methods focus on entire scene reconstruction, some methods~\cite{Zhi2021semanticnerf, wu2022object, Wu2023objectsdfplus} focus on individual objects in the scene. 
Semantic NeRF \cite{Zhi2021semanticnerf} uses a semantic head to predict labels for each position, supervised by the segmentation mask.
A similar idea is also used in \cite{huang2022hhor}, where the method predicts a label for each position and then uses it to separate the SDF of the human and the object. 
However, this approach---while recovering correct labels at the surface---produces incorrect labels \textit{within} the surface, making it feasible only for minimal occlusion and contact. 
The approach closest to ours is  ObjectSDF++~\cite{Wu2023objectsdfplus}, which predicts different SDFs for each object in the scene. 
Whereas they supervise the SDF separation using only opacity, we use the separately rendered colour of the two objects for supervision. 
Importantly, even though ObjectSDF++ proposes an extra SDF distinction regularisation term, it does not guarantee non-penetrating geometries. 
In contrast, we use an opacity regularisation term that incentivises the two SDFs to have disjoint opacities, thereby resulting in no (in most cases) or minimal interpenetration. 
Please refer to ~\cref{sec:Joint_SDF} for more details. 

\paragraph{Human-Object Interaction}

\begin{figure*}
    \centering
    \hspace*{10pt}\includegraphics[width=1.01\linewidth]{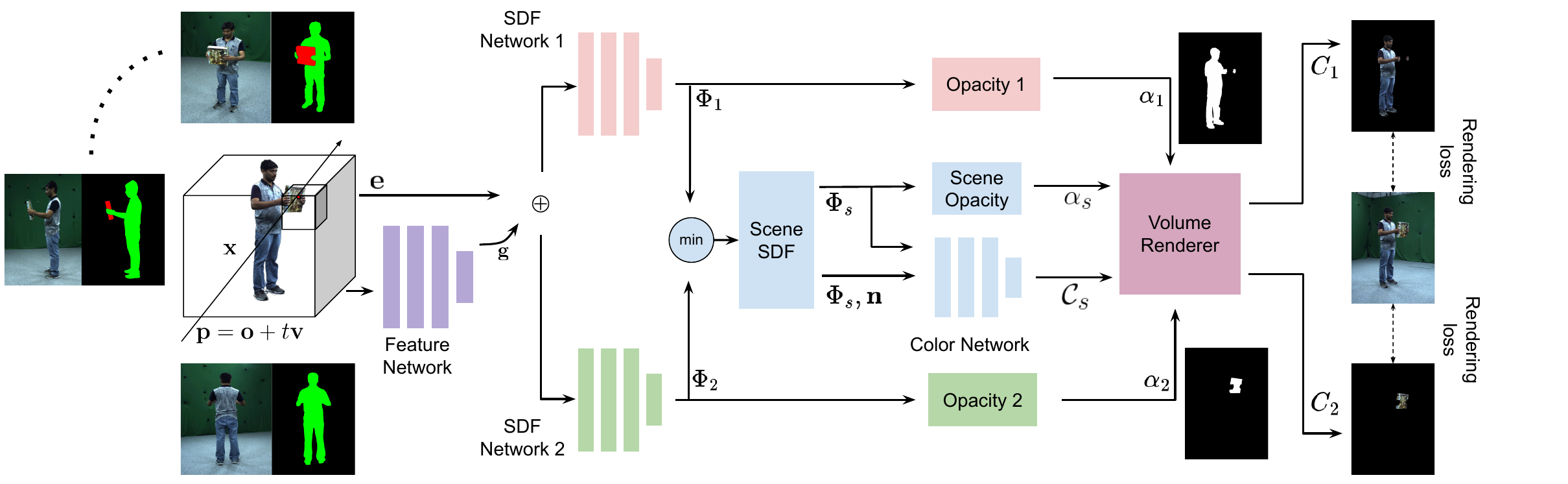}
    \caption{\textbf{Schematic overview of our framework.} 
    We semantically segment the input multi-view images into the background and the areas corresponding to two interacting objects. 
    The scene is encoded using a shared, multi-resolution hash grid encoding $\mathbf{e}$ and 
    the shared features are decoded using two separate SDF MLPs to produce corresponding SDFs $\Phi_1$ and $\Phi_2$.
    The per-point colour $\mathcal{C}_s$ is estimated from the joint scene SDF  composed using $\Phi_s = \Phi_1 \cup \Phi_2$.
    Finally, we integrate the colours of the sampled points in the ray by $\alpha$-blending  the individual opacities, $\alpha_1$ and $\alpha_2$, ensuring clean separation boundaries between the two (see \cref{eq:color_compositing}). 
    The entire framework is supervised using the rendering loss and additional regularisers (see~\cref{eq:total_loss}).
    }
    \label{fig:joint_sdf}
    \vspace{-1em}
\end{figure*}
A widely arising scenario is human-object interaction, which our method can also handle since it is applicable to arbitrary objects. 
Human-object interaction has been extensively studied in the literature. 
While some previous works \cite{kwon2021h2o,hampali2020honnotate,Brahmbhatt2020contactpose,chao2021dexycb,Freihand2019,FirstPersonAction_CVPR2018,DecafTOG2023} focus only on hands interacting with objects, 
many of the recent works \cite{bhatnagar22behave,GRAB:2020,jiang2022fullbody,xie2022chore,fan2023arctic,huang2022intercap,zhang2020phosa,Li_3DV2022,zhang2023neuraldome,jiang2022neuralhofusion,tretschk2024scenerflow} consider whole body interacting with the object.
In many scenarios of humans represented by entire bodies interacting with objects, the latter are often substantially smaller than humans. 
Methods falling into this category can be broadly divided into two groups: Methods based on template fitting or multi-view reconstruction methods. 
For template fitting, most methods, leverage SMPL or SMPL-X \cite{SMPL:2015,SMPL-X:2019}, 
along with a pre-acquired template of the object to fit marker-based motion capture data \cite{GRAB:2020,fan2023arctic}
or (sparse) multi-view RGB-D data \cite{bhatnagar22behave, huang2022intercap}. 
Some methods even attempt to fit templates to a single RGB image \cite{zhang2020phosa,xie2022chore}.
On the other hand, multi-view reconstruction methods, most relevant to our work, 
can recover accurate geometry and high-quality textures using multi-view RGB(D) images. 
Similarly to our method, NeuralDome~\cite{zhang2023neuraldome} and Neural-HOFusion~\cite{jiang2022neuralhofusion} use segmentation masks to separate humans and objects from multi-view images. 
However, in contrast to our work, both use a layer-wise NeRF representation that reconstructs humans and objects in isolation and then fuses them to recover final outputs. 
In contrast, our formulation uses a unified density and color field for the whole scene, but decodes separate geometries (as Signed Distance Fields) for the human and object. 
NeuralDome \cite{zhang2023neuraldome} uses SMPL-X and object template as prior for tracking. 
NeuralHOFusion \cite{jiang2022neuralhofusion} also uses object templates for better reconstruction. 
Note that our method does not require any 3D object templates free per default (the supplement discusses the case when one is available). 

\section{Background}
Our method is based on an implicit surface representation and uses volumetric rendering for image supervision.
It, therefore, builds on top of existing surface representation and volume rendering methods like NeuS~\cite{wang2021neus} 
and Instant-NGP~\cite{mueller2022instant}. 
We briefly discuss them below. 
\subsection{Neural Implicit Surfaces} 
NeuS \cite{wang2021neus} 
is an implicit multi-view reconstruction method that extends NeRF \cite{mildenhall2020nerf} by representing the surface and appearance of a scene as a Signed Distance Function (SDF),  $\Phi(\mathbf{x}): \mathbb{R}^3 \rightarrow \mathbb{R}$ and a radiance field $c(\mathbf{x}, \mathbf{v}): \mathbb{R}^5 \rightarrow \mathbb{R}^3$,  respectively. 
The surface $\mathcal{S}$ is defined by the zero level set of the SDF, $\mathcal{S} = \{ \mathbf{x} \in \mathcal{R}^3 | \Phi(\mathbf{x}) = 0 \}$ and the pixel colour is obtained by volumetrically rendering the colours $\hat{C}$ along the ray,  $\mathbf{p} = \mathbf{o} + i \cdot \mathbf{v}$, shot from the camera's origin $\mathbf{o}$ in the direction $\mathbf{v}$, through pixel $p$ using the rendering equation: 
\begin{equation} \label{eq:vol_render}
    \mathbf{\hat{C}}(p) = \sum_i^{N} T_i \alpha_i c_i, 
\end{equation}
where $T_i$ is the accumulated transmittance, $\alpha_i$ is the opacity and $c_i$ is the colour of $i^{th}$ sample along the ray. 
$\alpha_i$ and 
$T_i$ are obtained directly from 
$\Phi$ 
using 
\begin{equation}\label{eq:neus_opacity}
    \alpha_i = \max\left(\frac{\sigma(\Phi_i) - \sigma(\Phi_{i+1})}{\sigma(\Phi_i))}, 0 \right), 
\end{equation}
where $\sigma(\Phi_i) = (1 + e^{-\beta \Phi_i})^{-1}$ is the sigmoid function and $\beta$ is a learnable parameter.

We extend this formulation 
for two SDFs and, thereby, opacities of two interacting objects (see \cref{sec:Method}). 

\subsection{Hashgrid Encoding}
Training NeuS in the originally proposed fashion is slow and requires hours. 
Recent works~\cite{mueller2022instant, zhao2022instantnsr, neus2, rosu2023permutosdf} accelerate it using 
multi-resolution hash grid encoding of 3D points. 
Müller \textit{et al.}~\cite{mueller2022instant} as first proposed a multi-resolution voxel grid such that the grids of different resolutions are represented by a hash table that maps a 3D point $\mathbf{x}$ to a learnable feature vector $h_l(\mathbf{x})$, with $l$ being the resolution level. 
All the feature vectors are concatenated to obtain the hash-encoded feature as $h(\mathbf{x}) = \{h_1(\mathbf{x}), h_2(\mathbf{x}), \ldots, h_L(\mathbf{x})\}$, where $L$ is the number of resolution levels. 
This representation (along with the CUDA implementation), speeds up the training substantially by three orders of magnitude and has been used to accelerate several surface reconstruction methods \cite{neus2, zhao2022instantnsr}.
We also adopt it in our method.

\section{Method} 
\label{sec:Method} 
Our goal is to separately recover the 3D geometry and the appearance of each object in the scene from multiple RGB views. 
With close interactions (leading to severe mutual occlusions), the key challenge is to recover a clean surface boundary while preventing inter-penetrations. 
We address the problem with a new neural approach illustrated in \cref{fig:joint_sdf}.
The scene observed from multiple views is first encoded using a shared multi-resolution hash grid. 
In the second step, the scene features are decoded as two SDFs using two MLP heads, one for each interacting object (\cref{sec:Joint_SDF}). 
Next, the individual opacity of each object, the overall scene opacity and the scene colour for each point in the ray are forwarded to the volume renderer that renders the images. 
The separation boundaries between different objects are obtained using ray colour integration with $\alpha$-blending (\cref{sec:losses}). 
We next discuss each step in detail. 
\subsection{Scene Representation}
\label{sec:Joint_SDF}

We are given a set of $K$ calibrated multi-view images $\mathcal{I} = \{I_i\}_{i=1}^K$ capturing two interacting objects (subscripted with $1$ and $2$) along with their corresponding segmentation masks $\mathcal{M}_1 = \{M_{1}^{1}, M_{1}^{2}, \hdots,$ $ M_{1}^{K}\}$ and $\mathcal{M}_2 = \{M_{2}^{1}, M_{2}^{2} \hdots, M_{2}^{K}\}$.
One can recover the foreground mask as their union: $\mathcal{M} = \mathcal{M}_1 \cup \mathcal{M}_2$. 
A na\"ive way to perform 3D reconstruction would be to use the set of masks corresponding to each object in isolation in an attempt to recover the corresponding surfaces using a multi-view reconstruction method such as NeuS~\cite{wang2021neus} or VolSDF \cite{yariv2021volume}. 
Unfortunately, such a solution is sub-optimal as it does not jointly account for all the densities, resulting in large gaps due to occlusion and poor separation boundary as shown in \cref{fig:naive_sdf} and \cref{table:human_object_geometric}. 
\par
Hence, our approach uses a \textit{shared} hash-encoding for both objects in the scene. 
For any 3D point $\mathbf{x}$---encoded using a \textit{shared} multi-resolution hashgrid as $\mathbf{e} = (\mathbf{x}, h(\mathbf{x}))$ 
---we estimate the signed distance to the two objects in the scene using two separate SDFs, $\Phi_1$ and $\Phi_2$, respectively, parameterised using two separate MLP heads. 
As shown in~\cref{fig:joint_sdf}, the feature extraction network provides hashgrid features $\mathbf{g} \in \mathbb{R}^{d_g}$ for each encoded position $\mathbf{e}$.
These features, along with the hashgrid encodings are the input to the SDF MLPs that produce $\Phi_1(\mathbf{e}, \mathbf{g})$ and $\Phi_2(\mathbf{e}, \mathbf{g})$.
The SDF of the joint scene, $\Phi_s(\mathbf{e}, \mathbf{g})$, can now be composed using:
\begin{equation}
\label{eq:union_sdf}
    \Phi_s(\mathbf{e}, \mathbf{g}) = \min(\Phi_1(\mathbf{e}, \mathbf{g}), \Phi_2(\mathbf{e}, \mathbf{g})). 
\end{equation}
\par
We also recover the colour of the scene, $\mathcal{C}_s(\mathbf{x}, \mathbf{v}, \mathbf{n}, \Phi_s, \mathbf{g}): \mathbb{R}^{38} \rightarrow \mathbb{R}^3$, which can be encoded as an MLP and conditioned on the position $\mathbf{x}\in \mathbb{R}^{3}$, spherical-harmonic encoded view direction $\mathbf{v}\in \mathbb{R}^{16}$, the surface normal $\mathbf{n}\in \mathbb{R}^{3}$, SDF value $\Phi_s\in \mathbb{R}$ and the hashgrid features $\mathbf{g} \in \mathbb{R}^{15}$. 
For brevity, we denote this colour MLP as $\mathcal{C}_s(\mathbf{x}, \mathbf{v})$ in future sections. 
Using the scene SDF, $\Phi_s$, and the colour values, $\mathcal{C}_s$, one can apply the volume rendering proposed in~\cite{wang2021neus} to render the scene for each camera pose and each object separately, which is
visualised in our supplementary video. 
\subsection{Interaction-aware Training} 
\label{sec:losses}
Given the per-object segmentation masks, we optimise the scene parameters defined above using the following loss formulation. 
Let $\mathbf{C}$ represent the segmented foreground ground-truth colour, while 
$\mathbf{C}_1 = \mathbf{C} \circ \mathcal{M}_1$ and $\mathbf{C}_2 = \mathbf{C} \circ \mathcal{M}_2$ be the segmented objects' ground-truth colours, where ``$\circ$'' indicates the Hadamard product. 
The rendering loss function $\mathcal{L}_{\text{color}}$ is then defined as: 
\begin{equation} \label{eq:partial_color_loss}
    \mathcal{\hat{L}}_{\text{color}} = \sum_p|\mathbf{\hat{C}}_1(\mathbf{p}) - \mathbf{C}_1(\mathbf{p})|_s + \sum_p |\mathbf{\hat{C}}_2(\mathbf{p}) - \mathbf{C}_2(\mathbf{p})|_s, 
\end{equation}
where $|\cdot|_s$ denotes the  Smooth-$\operatorname{L1}$ loss. 
\par
\noindent \textbf{Training Stabilisation:}
In practice, since the individual objects can be relatively smaller than the overall scene scale, using only these segmented colours for SDF supervision makes the training unstable, especially in the beginning.
To stabilise the training, especially in the earlier stages, we use the entire scene colour for supervision as well, with predicted scene colour calculated using~\cref{eq:vol_render}.
The modified final loss now reads as:
\begin{align} 
    \mathcal{L}_{\text{color}} = \hat{\mathcal{L}}_{\text{color}} + \sum_p |\mathbf{\hat{C}}_s(\mathbf{p}) - \mathbf{C}(\mathbf{p})|_s. 
    \label{eq:color_loss} 
\end{align} 
\par 
While the estimated scene colour $\mathcal{C}_s(\mathbf{x}, \mathbf{v})$ can be directly supervised with the RGB colour loss, it is insufficient to enforce that the learned object and the human SDFs are \textit{separate}. 
Hence, the next question is how to ensure that the colour loss leads to separation between the two SDFs without arbitrarily entangling the two geometries.
Towards this goal, we introduce an $\alpha$-blending colour loss and a regularisation term that constrains the opacities of the two fields. 
Recall that we construct the scene SDF $\Phi_{s}(\mathbf{e})$ as a union of the individual object SDFs, $\Phi_1(\mathbf{e})$ and $\Phi_2(\mathbf{e})$, as in~\cref{eq:union_sdf}. 
We can, therefore, recover the opacity of the individual objects, $\alpha_1^i$ and $\alpha_2^i$, at position $i$ using the respective SDFs (as in ~\cref{eq:neus_opacity}). 
Now, to recover the joint scene opacity $\alpha_s^i$, we $\alpha$-composite the opacity contributions from both $\alpha_1^i$ and $\alpha_2^i$: 
\begin{equation} \label{eq:alpha_compositing}
    \alpha^{i}_{s} = \alpha^{i}_{1} + \alpha^{i}_{2} - \alpha^{i}_{1} \alpha^{i}_{2}. 
\end{equation}
After substituting $\alpha^{i}_{s}$ from~\cref{eq:alpha_compositing} in the rendering equation \eqref{eq:vol_render}, we obtain: 
\begin{align}
\mathbf{\hat{C}_{s}}(\mathbf{p}) &= \sum_{i=1}^{N} T^{i}_{s} \left( \alpha^{i}_{1} + \alpha^{i}_{2} - \alpha^{i}_{1} \alpha^{i}_{2} \right) c^{i}_{s}  \nonumber \\
&= \sum_{i=1}^{N} T^{i}_{s} \alpha^{i}_{1} c^{i}_{s} + T^{i}_{s} \alpha^{i}_{2} c^{i}_{s} -  T^{i}_{s} \alpha^{i}_{1} \alpha^{i}_{2} c^{i}_{s}. 
\label{eq:color_compositing}
\end{align}
Here, the first two terms,  $\mathbf{\hat{C}}_1(\mathbf{p}) = \sum_{i=1}^{N} T^{i}_{s} \alpha^{i}_{1} c^{i}_{s}$ and $\mathbf{\hat{C}}_2(\mathbf{p}) = \sum_{i=1}^{N} T^{i}_{s} \alpha^{i}_{2} c^{i}_{s}$, represent the visible part of the two objects. 
Note, however, that the transmittance $T_s^i$ and colour $c_s^i$ terms correspond to the entire scene, and $T_s^i$ reaches close to 0, when obstructed by either of the objects, thus ensuring that the final colour output is occlusion-aware.
\par 
\textbf{Alpha-Blending Regularisation.}
To achieve separable reconstruction, we assume that all the objects in a scene are opaque.
Therefore, at any point, at least one of the two opacities, ($\alpha_1^i, \alpha_2^i$), should be $0$, such that.~$\alpha_1^i \alpha_2^i = 0$ in~\cref{eq:color_compositing}. 
This observation is key to ensuring clean separation boundaries between the different objects.
However, we cannot \textit{explicitly} enforce this constraint as there is no way to know which of the two opacities should be $0$.
Thus, we introduce the following $\alpha$-regularisation which ensures that each position is opaque due to the influence of only one of the two SDFs, thereby preventing SDF penetration: 
\begin{equation}
\label{eq:alpha_reg}
    \mathcal{L}_{\text{alpha}} = \sum_p \left(\exp\bigg(\frac{\beta}{\lambda_t} \cdot \alpha_1(\mathbf{p}) \cdot \alpha_2(\mathbf{p})\bigg) - 1\right), 
\end{equation}
where $\beta$ is the learnable parameter from~\cref{eq:neus_opacity}, $\lambda_t$ is a hyperparameter controlling the temperature of the exponential curve and $\alpha_1, \alpha_2 \geq 0$. 
Here, $\beta$ increases as the training converges, thereby regularising more for overlapping opacities at the later stages of training. 
We empirically find that the above-proposed $\alpha$-regularisation performs the best and show an ablation in \cref{table:ablations}. 
Finally, we employ the commonly used Eikonal regularisation term 
to obtain the correct SDFs: 
\begin{align} \label{eq:eikonal_reg}
    \mathcal{L}_{\text{eik}} = &\sum_\mathbf{x} \norm{ \nabla_\mathbf{x} \Phi_1(\mathbf{e}, \mathbf{g}) - 1 }^2 + \norm{\nabla_\mathbf{x} \Phi_2(\mathbf{e}, \mathbf{g}) - 1}^2 + \nonumber \\ 
     & + \norm{\nabla_\mathbf{x} \Phi_s(\mathbf{e}, \mathbf{g}) - 1}^2. 
\end{align}
The resulting total loss can now be written as:
\begin{equation} \label{eq:total_loss}
    \mathcal{L}_{\text{recon}} = \mathcal{L}_{\text{color}} + \lambda_\alpha \mathcal{L}_{\text{alpha}} + \lambda_{\text{eik}} \mathcal{L}_{\text{eik}}. 
\end{equation}
We use $\lambda_{\alpha} = 0.1$, $\lambda_{\text{eik}} = 0.01$ and $\lambda_t = 100.0$ as hyperparameters in all our experiments. 
\par

\noindent \textbf{Opacity \textit{vs.}~direct SDF regularisation.} 
While we 
ensure separability by regularising the per-object opacities, another possible approach would be to 
regularise at the SDF level: 
Specifically, one could enforce both SDFs $\Phi_1$ and $\Phi_2$ to be not negative at the same point. 
We empirically observe that the proposed $\alpha$-regularisation performs best and show the corresponding ablation in~\cref{table:ablations}. 

\section{Experiments}

\par
\noindent 
\textbf{Metrics:} We report novel-view synthesis evaluations on commonly used metrics such as peak signal-to-noise ratio (PSNR), structural similarity index (SSIM) and learned perceptual image patch similarity (LPIPS).
We use bi-directional Chamfer distance to evaluate the 3D reconstruction quality. 
\par
As our method is agnostic to the object types, we demonstrate its effectiveness on four kinds of interactions: Human-object, hand-object, and object-object and human-human interaction. 
For the hand-object and object-object scenarios, we use scenes from the AffordPose~\cite{affordpose} and WildRGB-D~\cite{xia2024rgbd} datasets, respectively.
AffordPose is a synthetic dataset with ground-truth geometry. 
We use these ground-truth meshes to generate a synthetic multi-view dataset of 60 views for six objects. 
Human-object interaction is especially challenging due to the large differences between the scales of the two entities. 
To evaluate our method comprehensively, we capture and record a new dataset with various human-object interactions with objects of different scales and complexity (further details in supplementary \cref{sec:human_object_dataset}) and also evaluate our method on a few scenes of human-human interaction from ReMoCap\cite{ghosh2024remos} dataset.
The evaluation datasets also differ in the relative scales of the objects in the scene. 
For example, the human-object evaluation dataset consists of small scale objects in a large capture dome, as opposed to the Wild-RGBD dataset.
\par 
\noindent \textbf{Comparisons:} We compare our method against ObjectSDF++~\cite{Wu2023objectsdfplus} and ``Segmented Neus2''. 
As NeuS2~\cite{neus2} is not designed to be instance-specific, we train it separately for each object in the scene by providing the corresponding segmentation masks (henceforth referred to as ``Segmented NeuS2''). 
ObjectSDF++, on the other hand, is a state-of-the-art method that reconstructs objects in a scene separately.%

For human-object, human-human and object-object datasets, we evaluate 3D reconstruction quality for the overall scene. 
As we do not have a ground truth here, we consider a NeuS2~\cite{neus2} model trained on the entire scene (agnostic of the objects) as the pseudo ground truth. 
This allows us to compare the compositionally reconstructed scene with the non-compositional reconstruction of the scene, which can be treated as an upper bound on the reconstruction quality.
In the case of human-object interaction, we also evaluate the object reconstruction separately with respect to 3D scanned templates--we first align the pre-scanned 3D object shape with the reconstructed mesh (extracted using marching cubes) using the rigid ICP~\cite{icp} algorithm and then compute the Chamfer distance. 
We use the ground-truth meshes provided in AffordPose to compute the Chamfer distance metric for both the hand and the object separately.

\subsection{Geometric Evaluation}
\begin{figure}
    \centering \includegraphics[width=1.0\linewidth]{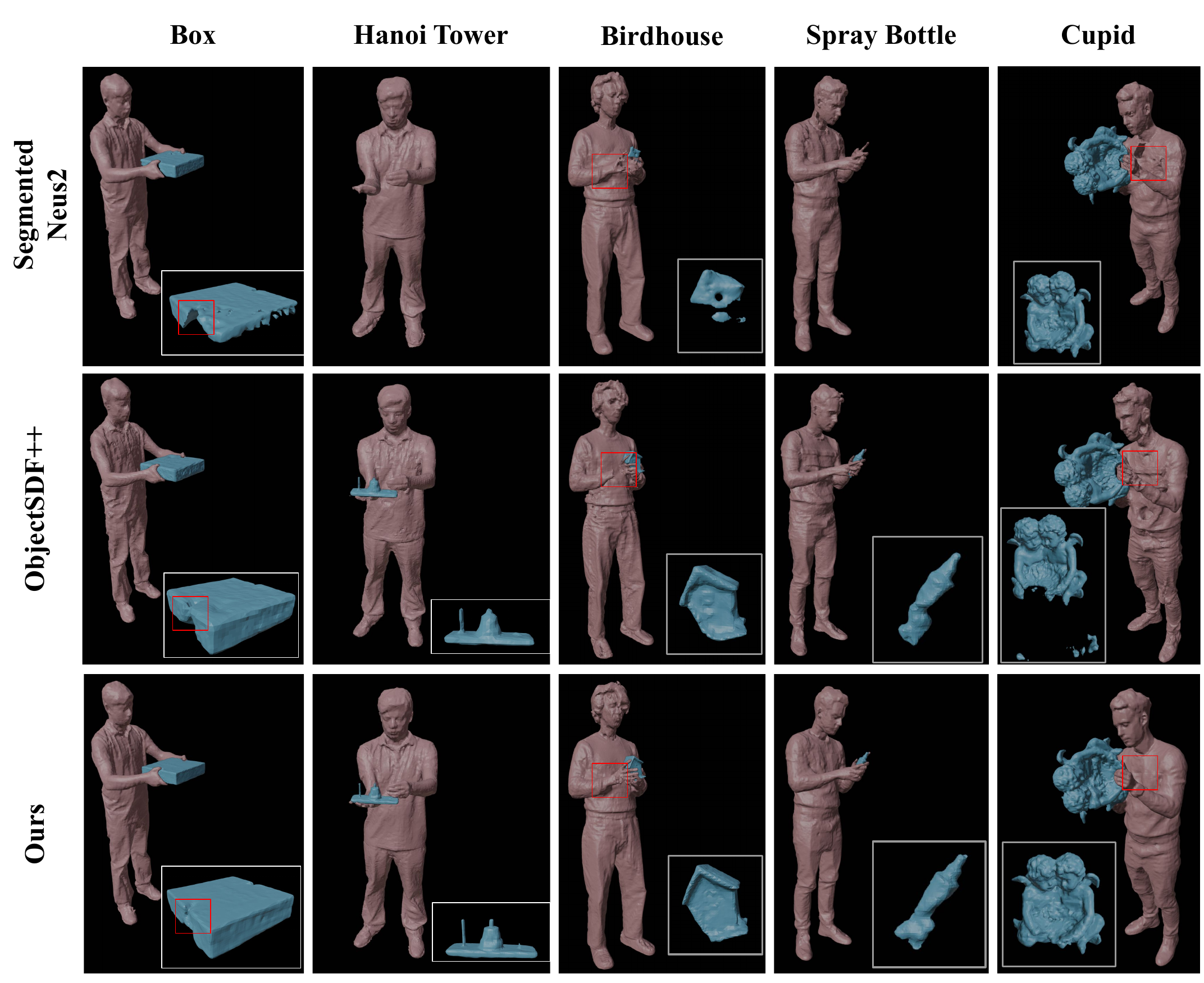}
    \caption{Qualitative comparison of the reconstructed geometry. In most scenes, we obtain better geometry, with fewer deformations near the contact regions. Best viewed when zoomed.
    }
    \label{fig:human_object_geometry}
    \vspace{-1em}
\end{figure}
\begin{figure}
    \centering
    \includegraphics[width=\linewidth]{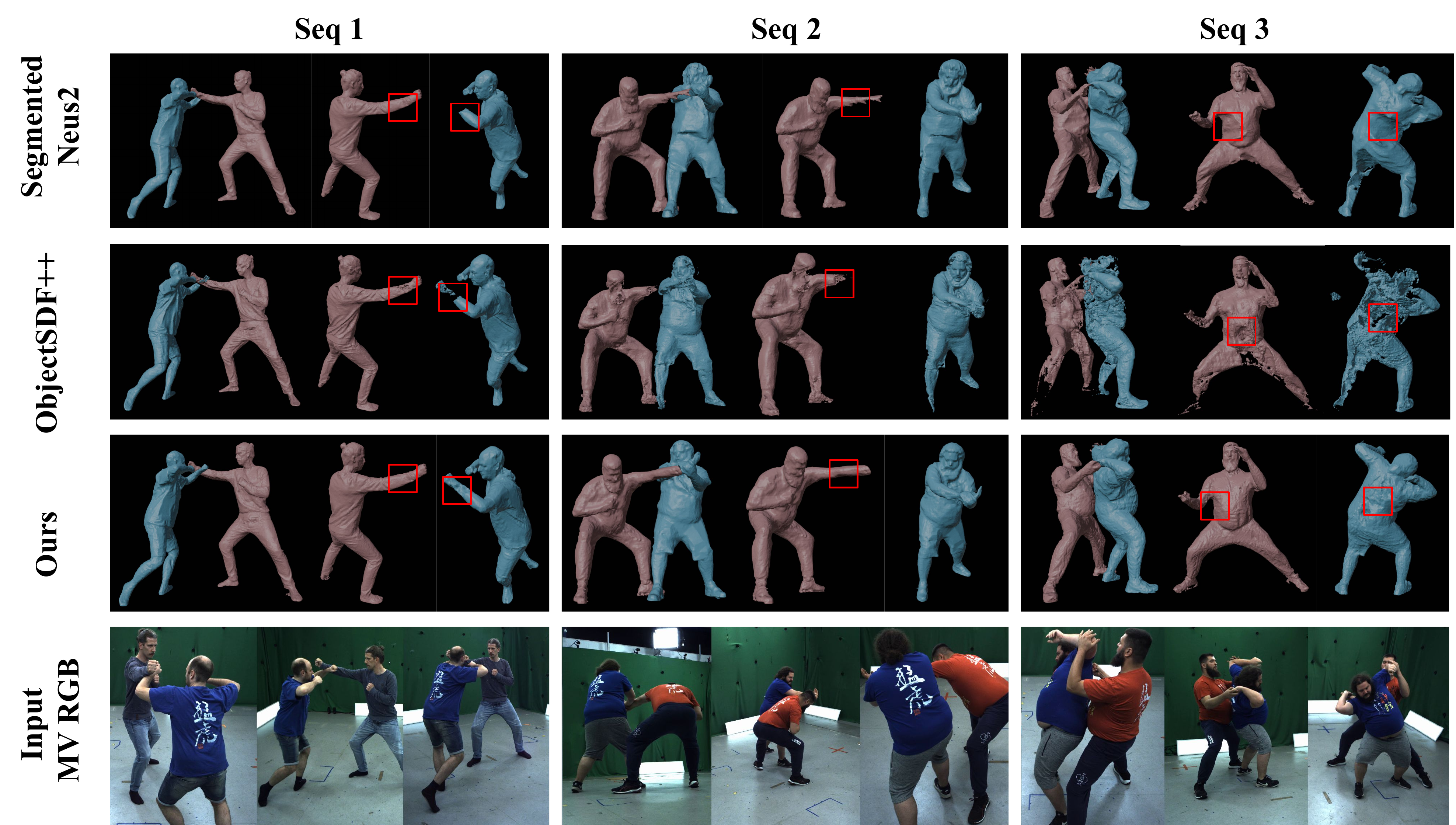}
    \caption{Qualitative comparison of 3D scene reconstructions with human-human interaction along with selected multi-view (MV) input images. 
    Digital zoom recommended. 
    } 
    \label{fig:human_human_geometry}
    \vspace{-1em}
\end{figure}
\begin{figure}
    \centering    
    \includegraphics[width=\linewidth]{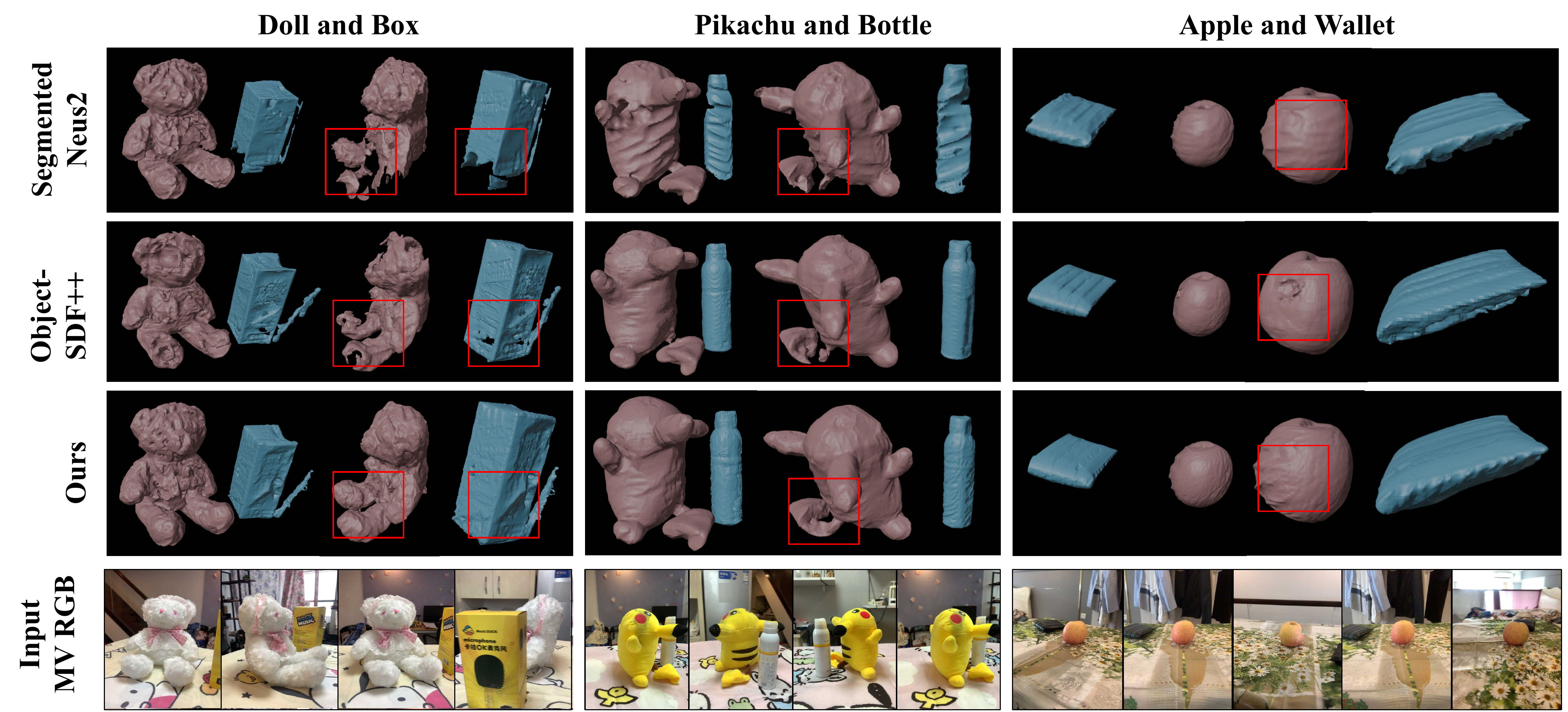}
    \caption{Qualitative comparison of reconstruction of scenes involving two objects in proximity, along with samples from the multi-view (MV) input images. 
    Digital zoom recommended. 
    } \label{fig:object_object_geometry}
    \vspace{-1em}
\end{figure}

\begin{table}[t]
\resizebox{\linewidth}{!} {
    \begin{tabular}{|l|c|cc|cc|cc|}
\hline
\textbf{Scene}                         & \textbf{Seq ID} & \multicolumn{2}{c|}{\textbf{Segmented NeuS2}}                 & \multicolumn{2}{c|}{\textbf{ObjectSDF++}}                     & \multicolumn{2}{c|}{\textbf{Ours}}                            \\ \hline
                                       & \textbf{}       & \multicolumn{1}{c|}{\textbf{Overall scene}} & \textbf{Object} & \multicolumn{1}{c|}{\textbf{Overall scene}} & \textbf{Object} & \multicolumn{1}{c|}{\textbf{Overall scene}} & \textbf{Object} \\ \hline
\multirow{2}{*}{\textbf{Box}}          & 1               & \multicolumn{1}{c|}{\textbf{4.86}}          & 26.06           & \multicolumn{1}{c|}{11.08}                  & 10.14           & \multicolumn{1}{c|}{5.65}                   & \textbf{7.9}    \\ \cline{2-8} 
                                       & 2               & \multicolumn{1}{c|}{\textbf{11.18}}         & 42.87           & \multicolumn{1}{c|}{18.96}                  & \textbf{12.62}  & \multicolumn{1}{c|}{11.41}                  & 38.46           \\ \hline
\multirow{2}{*}{\textbf{Book}}         & 1               & \multicolumn{1}{c|}{6.78}                   & -               & \multicolumn{1}{c|}{13.31}                  & 9.13            & \multicolumn{1}{c|}{\textbf{5.35}}          & \textbf{7.33}   \\ \cline{2-8} 
                                       & 2               & \multicolumn{1}{c|}{7.30}                   & -               & \multicolumn{1}{c|}{11.96}                  & 9.35            & \multicolumn{1}{c|}{\textbf{5.95}}          & \textbf{8.28}   \\ \hline
\multirow{2}{*}{\textbf{Birdhouse}}    & 1               & \multicolumn{1}{c|}{4.70}                   & -               & \multicolumn{1}{c|}{9.27}                   & 13.2            & \multicolumn{1}{c|}{\textbf{4.50}}          & \textbf{10.99}  \\ \cline{2-8} 
                                       & 2               & \multicolumn{1}{c|}{4.96}                   & -               & \multicolumn{1}{c|}{8.74}                   & 14.33           & \multicolumn{1}{c|}{\textbf{4.38}}          & \textbf{11.13}  \\ \hline
\multirow{2}{*}{\textbf{Spray Bottle}} & 1               & \multicolumn{1}{c|}{\textbf{3.21}}          & -               & \multicolumn{1}{c|}{9.23}                   & 10.89           & \multicolumn{1}{c|}{4.40}                   & \textbf{7.72}   \\ \cline{2-8} 
                                       & 2               & \multicolumn{1}{c|}{\textbf{3.04}}          & -               & \multicolumn{1}{c|}{9.60}                   & 9.67            & \multicolumn{1}{c|}{4.19}                   & \textbf{7.48}   \\ \hline
\multirow{2}{*}{\textbf{Hanoi Tower}}  & 1               & \multicolumn{1}{c|}{\textbf{4.13}}          & -               & \multicolumn{1}{c|}{10.06}                  & 10.86           & \multicolumn{1}{c|}{4.30}                   & \textbf{8.59}   \\ \cline{2-8} 
                                       & 2               & \multicolumn{1}{c|}{\textbf{3.66}}          & -               & \multicolumn{1}{c|}{9.47}                   & 10.53           & \multicolumn{1}{c|}{4.30}                   & \textbf{8.64}   \\ \hline
\multirow{2}{*}{\textbf{Cupid}}        & 1               & \multicolumn{1}{c|}{8.11}                   & 119.20          & \multicolumn{1}{c|}{36.16}                  & 16.76           & \multicolumn{1}{c|}{\textbf{5.79}}          & \textbf{11.71}  \\ \cline{2-8} 
                                       & 2               & \multicolumn{1}{c|}{6.96}                   & 51.84           & \multicolumn{1}{c|}{18.81}                  & 63.34           & \multicolumn{1}{c|}{\textbf{5.92}}          & \textbf{8.66}   \\ \hline\hline
\textbf{Mean}        &                & \multicolumn{1}{c|}{5.74}                   & 59.99          & \multicolumn{1}{c|}{13.89}                  & 15.90           & \multicolumn{1}{c|}{\textbf{5.51}}          & \textbf{11.40} \\ \hline
\end{tabular}
}
    \caption{
    3D reconstruction accuracy of the \textit{overall scene} and individual \textit{object}, on human-object evaluation dataset, using Chamfer Distance (lower the better).
    }
    \label{table:human_object_geometric}
\end{table}

\textbf{Human-Object Interaction:} Table~\ref{table:human_object_geometric} tabulates the quantitative comparisons of our method with ObjectSDF++ and Segmented Neus2 for the whole scenes and also individual objects separately. 
While we outperform ObjectSDF++ on most scenes, an interesting pattern emerges: 
The \textit{Overall scene} in ~\cref{table:human_object_geometric} shows Segmented NeuS2 achieves consistently lower Chamfer distance for the scenes involving small objects like \textit{Spray Bottle} and \textit{Hanoi Tower}. 
The full-scene results are dominated by the reconstruction of the human. 
However, these results deteriorate 
with a relatively larger object like \textit{Cupid}. 
As the occlusions on the human body grow (due to the larger object size), the Segmented NeuS2 struggles to maintain artefact-free reconstruction. 
Further results in the \textit{Object} in ~\cref{table:human_object_geometric} indicate that indeed, the Segmented NeuS2 does not recover the object geometries in most cases from the supervision using segmented objects alone.
We believe that this is because Neus2 tries to \textit{carve away} regions that are segmented out because of occlusion. This causes conflicting optimisation goals for different camera views depending on whether the object is visible. Since this can be a significant volume relative to the total volume for smaller objects, Neus2 fails to converge.
We show a qualitative comparison of our 3D reconstructions in~\cref{fig:human_object_geometry}.
\begin{table}[t]
\resizebox{\columnwidth}{!}{
\begin{tabular}{|l|ll|ll|ll|}
\hline
\multirow{2}{*}{\textbf{Scene}}    & \multicolumn{2}{c|}{\textbf{Segmented Neus2}}                   & \multicolumn{2}{c|}{\textbf{ObjectSDF++}}             & \multicolumn{2}{c|}{\textbf{Ours}}                    \\ \cline{2-7}
                  & \multicolumn{1}{l|}{\textbf{Hand}} & \textbf{Object} & \multicolumn{1}{l|}{\textbf{Hand}} & \textbf{Object} & \multicolumn{1}{l|}{\textbf{Hand}} & \textbf{Object} \\ \hline
\textbf{Bag}      & \multicolumn{1}{l|}{7.51}       & \gold{4.16}        & \multicolumn{1}{l|}{5.86}       & 11.30        & \multicolumn{1}{l|}{\gold{5.83}}       & 4.62        \\ \hline
\textbf{Bottle}   & \multicolumn{1}{l|}{8.28}       & 8.28        & \multicolumn{1}{l|}{6.28}       & 2.76        & \multicolumn{1}{l|}{\gold{5.40}}       & \gold{1.77}        \\ \hline
\textbf{Earphone} & \multicolumn{1}{l|}{7.18}               &  7.15 & \multicolumn{1}{l|}{5.85}               &  6.18               & \multicolumn{1}{l|}{\gold{5.44}}               & \gold{6.17}                \\ \hline
\textbf{Knife}    & \multicolumn{1}{l|}{6.06}       & 4.10        & \multicolumn{1}{l|}{\gold{5.64}}       & 4.32        & \multicolumn{1}{l|}{5.68}       & \gold{3.24}        \\ \hline
\textbf{Pot}      & \multicolumn{1}{l|}{-}              & 3.13        & \multicolumn{1}{l|}{\gold{6.94}}       & 11.97        & \multicolumn{1}{l|}{7.49}      & \gold{2.79}        \\ \hline
\textbf{Scissors} & \multicolumn{1}{l|}{6.00}       & 28.75        & \multicolumn{1}{l|}{6.52}       & 4.24        & \multicolumn{1}{l|}{\gold{5.42}}       & \gold{4.20}        \\ \hline\hline
\textbf{Mean} & \multicolumn{1}{l|}{7.00}       & 9.26        & \multicolumn{1}{l|}{6.18}       & 6.79        & \multicolumn{1}{l|}{\gold{5.87}}       & \gold{3.80}        \\ \hline
\end{tabular}
}
\caption{
3D reconstruction accuracy for different object sequences on the AffordPose dataset. Chamfer distance (lower the better) is calculated against the ground-truth meshes provided. Segmented Neus2 reconstruction fails for the hand in the \textit{Pot} scene, which is represented with ``-''.} 
\label{table:hand_object_geometry}
\vspace{-1em}
\end{table}
\par
\noindent \textbf{Hand-Object Interaction:} 
We show quantitative and qualitative comparisons in \cref{table:hand_object_geometry} and  \cref{fig:hand_object_geometry} in the supplement, respectively. 
ObjectSDF++ suffers from denting artefacts in the occluded areas, whereas our method generates a smoother surface. 
Both methods are at par when the object is only mildly in contact (as in the case of comparably thin scissors). 
\par
\noindent \textbf{Human-Human Interaction:} We show qualitative and quantitative comparisons in \cref{fig:human_human_geometry} 
and \cref{table:human_human_geometry}, respectively. 
Similarly to the previous section, the Segmented Neus2 reconstructs humans with missing sections, while reconstruction near contact areas in ObjectSDF++ shows severe artefacts. 
Note that for the case of Seq 1---even though numerically Segmented Neus2 appears to be slightly better---we can see in the qualitative results that our method reconstructs the hand near occlusions significantly better. 

\begin{table}[]
\resizebox{\columnwidth}{!}{
\begin{tabular}{|l|c|c|c|}
\hline
\textbf{Seq ID} & \multicolumn{1}{l|}{\textbf{Segmented Neus2}} & \multicolumn{1}{l|}{\textbf{ObjectSDF++}} & \multicolumn{1}{l|}{\textbf{Ours}} \\ \hline
\textbf{1}    & \gold{4.36}                                          & 25.05                                     & 4.73                               \\ \hline
\textbf{2}    & 9.8                                           & 21.14                                     & \gold{6.19}                               \\ \hline
\textbf{3}    & 13.71                                         & 43.26                                     & \gold{6.67}                               \\ \hline
\end{tabular}}
\vspace{-1em}
\caption{Comparison of 3D reconstruction quality for human-human interaction. Chamfer distance metric (lower the better) is calculated against the overall scene reconstructed using Neus2.} 
\label{table:human_human_geometry}
\vspace{-1em}
\end{table}
\par
\noindent \textbf{Object-Object Reconstruction:} The qualitative and quantitative results for two interesting objects are shown in \cref{fig:object_object_geometry} and \cref{table:object_object_geometric}, respectively. 
Our method excels in two cases out of three. 
\begin{table}[]
\resizebox{\columnwidth}{!}{
\begin{tabular}{|l|c|c|c|}
\hline
\textbf{}        & \multicolumn{1}{l|}{\textbf{Segmented Neus2}} & \multicolumn{1}{l|}{\textbf{ObjectSDF++}} & \multicolumn{1}{l|}{\textbf{Ours}} \\ \hline
\textbf{Doll and Box}    & 7.44                                         & 11.88                                    & \gold{4.44}                              \\ \hline
\textbf{Pikachu and Bottle} & 4.11                                       & 3.96                                     & \gold{2.99}                              \\ \hline
\textbf{Apple and Wallet}   & \gold{2.72}                                         & 6.28                                     & 3.83                              \\ \hline
\end{tabular}}
\vspace{-1em}
\caption{Comparison of the 3D reconstruction quality for overall scene, on the WilDRGBD dataset using Chamfer Distance metric (lower the better).}
\label{table:object_object_geometric}
\vspace{-1em}
\end{table}
\subsection{Appearance Evaluation}
\begin{figure}
    \centering    \includegraphics[width=\linewidth]{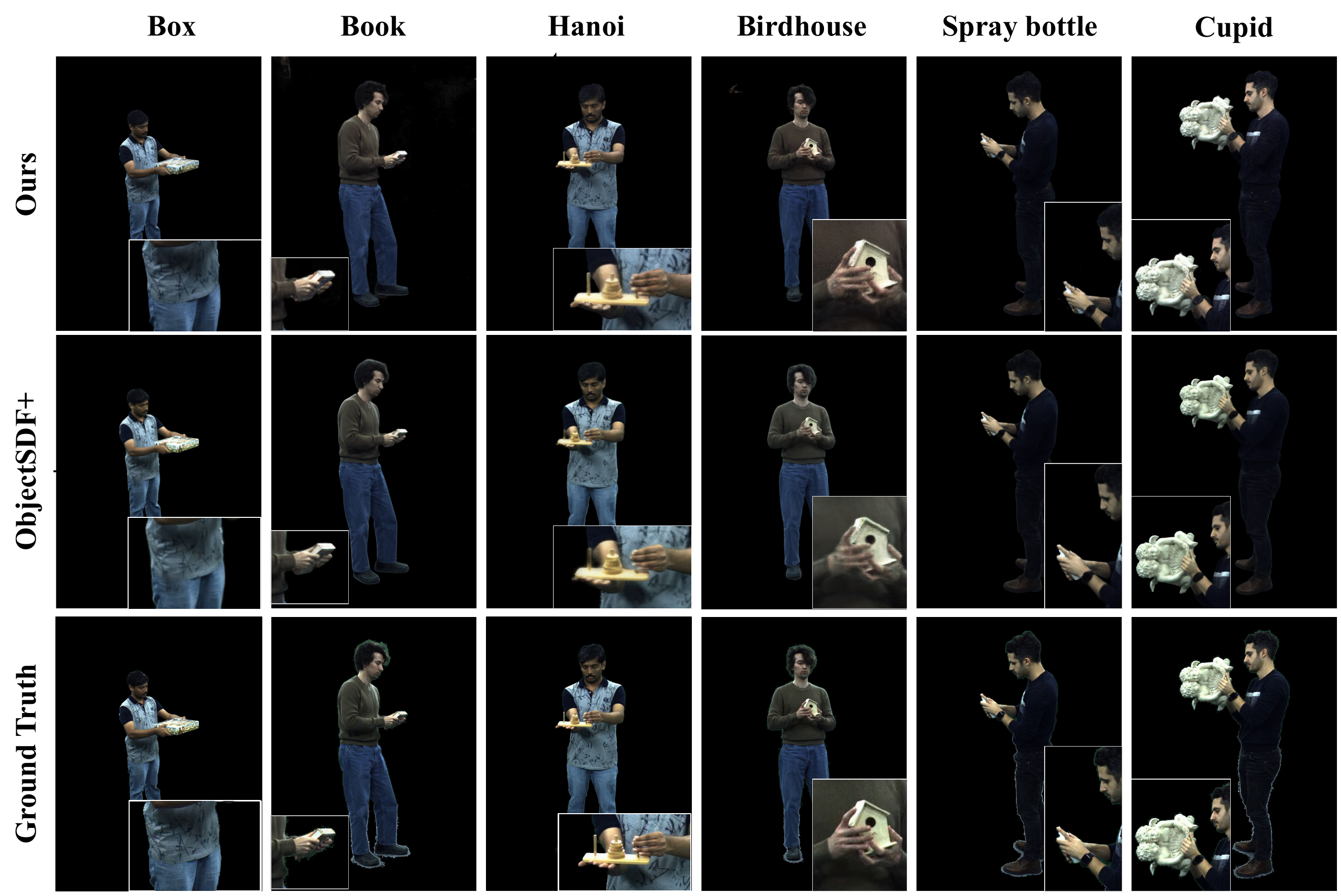}
    \caption{Qualitative comparison of novel view synthesis on human-object scenes. 
    The results of ObjectSDF++ are blurrier than our rendered views, especially around the object. 
    Digital zoom recommended. 
    } \label{fig:human_object_appearance}
    \vspace{-1em}
\end{figure}
To evaluate the quality of novel view synthesis, we render the entire scenes into a held-out set of views. 
\begin{table}[t]
\resizebox{\columnwidth}{!}{
\begin{tabular}{|l|l|ll|ll|ll|}
\hline
\textbf{Scene}          & \multicolumn{1}{c|}{\shortstack{\textbf{Seq}\\\textbf{ID}}} & \multicolumn{2}{c|}{\textbf{PSNR}$\uparrow$}                                             & \multicolumn{2}{c|}{\textbf{SSIM} $\uparrow$}                                             & \multicolumn{2}{c|}{\textbf{LPIPS} $\downarrow$}                                            \\ \hline
\textbf{}               & \multicolumn{1}{c|}{\textbf{}}       & \multicolumn{1}{c|}{\shortstack{\textbf{Object-}\\\textbf{SDF++}}} & \multicolumn{1}{c|}{\textbf{Ours}} & \multicolumn{1}{c|}{\shortstack{\textbf{Object-}\\\textbf{SDF++}}} & \multicolumn{1}{c|}{\textbf{Ours}} & \multicolumn{1}{c|}{\shortstack{\textbf{Object-}\\\textbf{SDF++}}} & \multicolumn{1}{c|}{\textbf{Ours}} \\ \hline
\multicolumn{1}{|l|}{\textbf{Box}}                             & \multicolumn{1}{c|}{1} & \multicolumn{1}{c|}{27.28} & \multicolumn{1}{c|}{\textbf{30.39}} & \multicolumn{1}{c|}{0.95}          & \multicolumn{1}{c|}{\textbf{0.96}} & \multicolumn{1}{c|}{0.08}          & \multicolumn{1}{c|}{\textbf{0.07}} \\ \hline
\multicolumn{1}{|l|}{\textbf{}}                                & \multicolumn{1}{c|}{2} & \multicolumn{1}{c|}{28.26} & \multicolumn{1}{c|}{\textbf{29.64}} & \multicolumn{1}{c|}{0.95}          & \multicolumn{1}{c|}{\textbf{0.96}} & \multicolumn{1}{c|}{0.08}          & \multicolumn{1}{c|}{\textbf{0.07}} \\ \hline
\multicolumn{1}{|l|}{\multirow{2}{*}{\textbf{Book}}}           & \multicolumn{1}{c|}{1} & \multicolumn{1}{c|}{25.61} & \multicolumn{1}{c|}{\textbf{33.06}} & \multicolumn{1}{c|}{0.95}          & \multicolumn{1}{c|}{\textbf{0.96}} & \multicolumn{1}{c|}{0.08}          & \multicolumn{1}{c|}{\textbf{0.06}} \\ \cline{2-8} 
\multicolumn{1}{|l|}{}                                         & \multicolumn{1}{c|}{2} & \multicolumn{1}{c|}{28.70} & \multicolumn{1}{c|}{\textbf{32.12}} & \multicolumn{1}{c|}{0.95}          & \multicolumn{1}{c|}{\textbf{0.96}} & \multicolumn{1}{c|}{0.08}          & \multicolumn{1}{c|}{\textbf{0.07}} \\ \hline
\multicolumn{1}{|l|}{\multirow{2}{*}{\textbf{Birdhouse}}}      & \multicolumn{1}{c|}{1} & \multicolumn{1}{c|}{29.82} & \multicolumn{1}{c|}{\textbf{33.71}} & \multicolumn{1}{c|}{0.96}          & \multicolumn{1}{c|}{\textbf{0.97}} & \multicolumn{1}{c|}{0.08}          & \multicolumn{1}{c|}{\textbf{0.05}} \\ \cline{2-8} 
\multicolumn{1}{|l|}{}                                         & \multicolumn{1}{c|}{2} & \multicolumn{1}{c|}{26.37} & \multicolumn{1}{c|}{\textbf{32.69}} & \multicolumn{1}{c|}{0.93}          & \multicolumn{1}{c|}{\textbf{0.96}} & \multicolumn{1}{c|}{0.10}          & \multicolumn{1}{c|}{\textbf{0.06}} \\ \hline
\multicolumn{1}{|l|}{\multirow{2}{*}{\textbf{\shortstack{Spray \\ Bottle}}}}   & \multicolumn{1}{c|}{1} & \multicolumn{1}{c|}{29.90} & \multicolumn{1}{c|}{\textbf{32.73}} & \multicolumn{1}{c|}{0.97}          & \multicolumn{1}{c|}{\textbf{0.98}} & \multicolumn{1}{c|}{0.07}          & \multicolumn{1}{c|}{\textbf{0.04}} \\ \cline{2-8} 
\multicolumn{1}{|l|}{}                                         & \multicolumn{1}{c|}{2} & \multicolumn{1}{c|}{26.32} & \multicolumn{1}{c|}{\textbf{36.38}} & \multicolumn{1}{c|}{0.94}          & \multicolumn{1}{c|}{\textbf{0.98}} & \multicolumn{1}{c|}{0.10}          & \multicolumn{1}{c|}{\textbf{0.04}} \\ \hline
\multicolumn{1}{|l|}{\multirow{2}{*}{\textbf{\shortstack{Hanoi \\ Tower}}}} & \multicolumn{1}{c|}{1} & \multicolumn{1}{c|}{27.25} & \multicolumn{1}{c|}{\textbf{30.36}} & \multicolumn{1}{c|}{0.95}          & \multicolumn{1}{c|}{\textbf{0.96}} & \multicolumn{1}{c|}{0.09}          & \multicolumn{1}{c|}{\textbf{0.07}} \\ \cline{2-8} 
\multicolumn{1}{|l|}{}                                         & \multicolumn{1}{c|}{2} & \multicolumn{1}{c|}{26.30} & \multicolumn{1}{c|}{\textbf{29.63}} & \multicolumn{1}{c|}{\textbf{0.96}} & \multicolumn{1}{c|}{\textbf{0.96}} & \multicolumn{1}{c|}{\textbf{0.07}} & \multicolumn{1}{c|}{\textbf{0.07}} \\ \hline
\multicolumn{1}{|l|}{\multirow{2}{*}{\textbf{Cupid}}}          & \multicolumn{1}{c|}{1} & \multicolumn{1}{c|}{29.92} & \multicolumn{1}{c|}{\textbf{34.10}} & \multicolumn{1}{c|}{0.96}          & \multicolumn{1}{c|}{\textbf{0.98}} & \multicolumn{1}{c|}{0.10}          & \multicolumn{1}{c|}{\textbf{0.05}} \\ \cline{2-8} 
\multicolumn{1}{|l|}{}                                         & \multicolumn{1}{c|}{2} & \multicolumn{1}{c|}{30.50} & \multicolumn{1}{c|}{\textbf{34.20}} & \multicolumn{1}{c|}{\textbf{0.97}} & \multicolumn{1}{c|}{\textbf{0.97}} & \multicolumn{1}{c|}{\textbf{0.08}} & \multicolumn{1}{c|}{\textbf{0.08}}          \\ \hline\hline
\multicolumn{1}{|l|}{\textbf{Mean}}                            & \multicolumn{1}{c|}{}  & \multicolumn{1}{c|}{28.02} & \multicolumn{1}{c|}{\textbf{32.42}} & \multicolumn{1}{c|}{0.95}          & \multicolumn{1}{c|}{\textbf{0.97}} & \multicolumn{1}{c|}{0.08}          & \multicolumn{1}{c|}{\textbf{0.06}} \\ \hline
\end{tabular}
}
\vspace{-1em}
\caption{Quantitative comparison of view synthesis on held-out views for the human-object dataset. 
We consistently outperform ObjectSDF++. 
}
\label{table:human_object_appearance}
\vspace{-1em}
\end{table}
We report the human-object novel-view results in~\cref{table:human_object_appearance}.
Again, we achieve consistently better performance than ObjectSDF++; see~\cref{fig:human_object_appearance} for the visualisations. 
One can observe blurring artefacts in ObjectSDF++ renderings, which are especially pronounced around the object.
As it supervises only the individual opacities, we hypothesize that the colour network of Wu \textit{et al.}~\cite{Wu2023objectsdfplus} assigns colours to any residual opacity, which is more likely to exist at the transition boundaries.
We also provide appearance evaluation for human-human interaction scenes in the supplementary \cref{sec:human_human_appearance_suppl}.

\subsection{Ablations}
\begin{figure}
    \centering
    \includegraphics[width=0.98\linewidth]{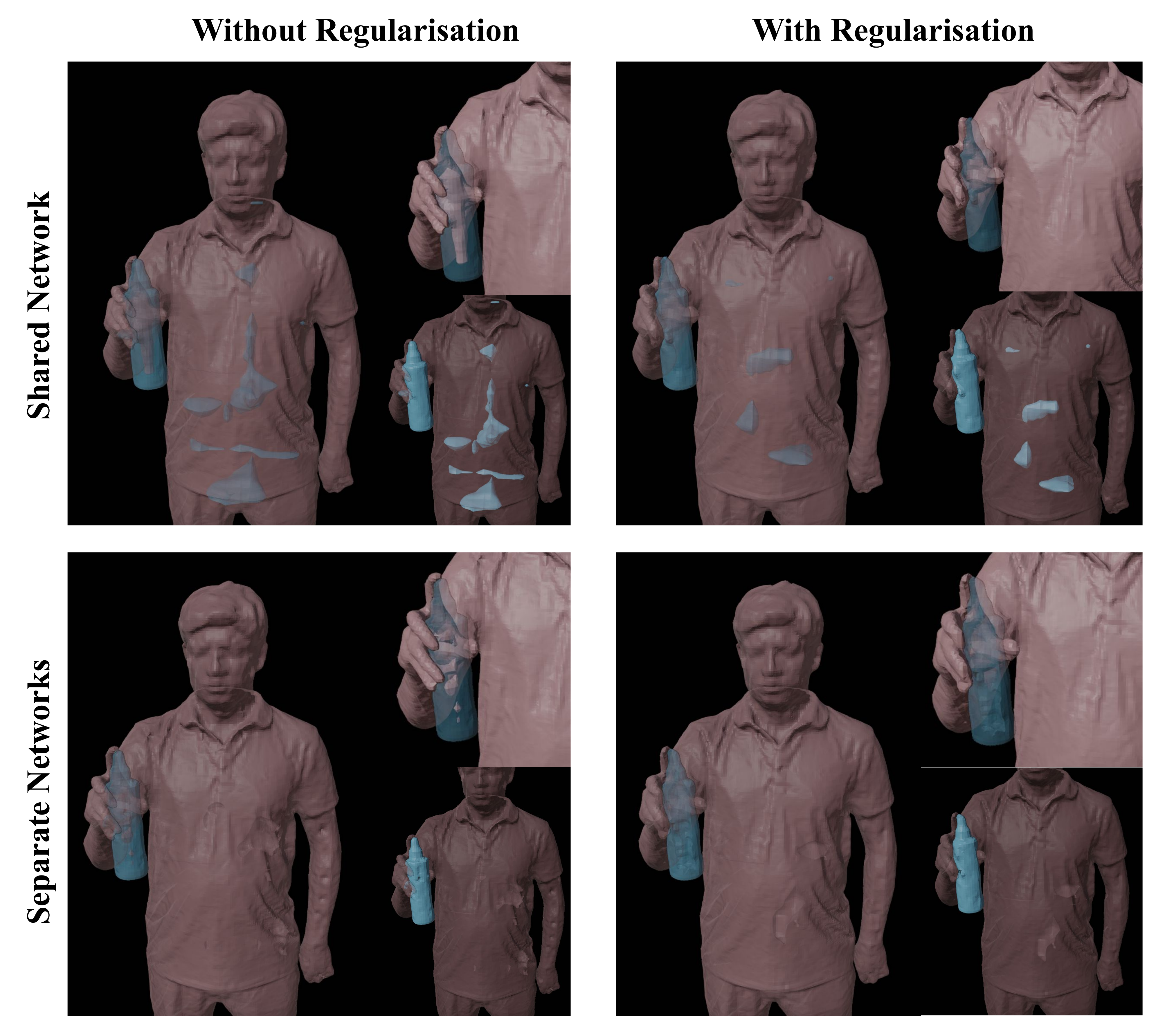}
    \caption[Ablations]{Qualitative comparison (ablations). We observe that the overall scene reconstruction largely remains the same, though the individual object and human reconstruction quality deteriorates because of phantom blobs formed underneath the surface (as highlighted inside the transparent surface) when we have a shared MLP or we do not use the alpha-regularisation.} 
    \label{fig:ablations}
\end{figure}

\begin{figure}
    \centering
    \includegraphics[width=\linewidth]{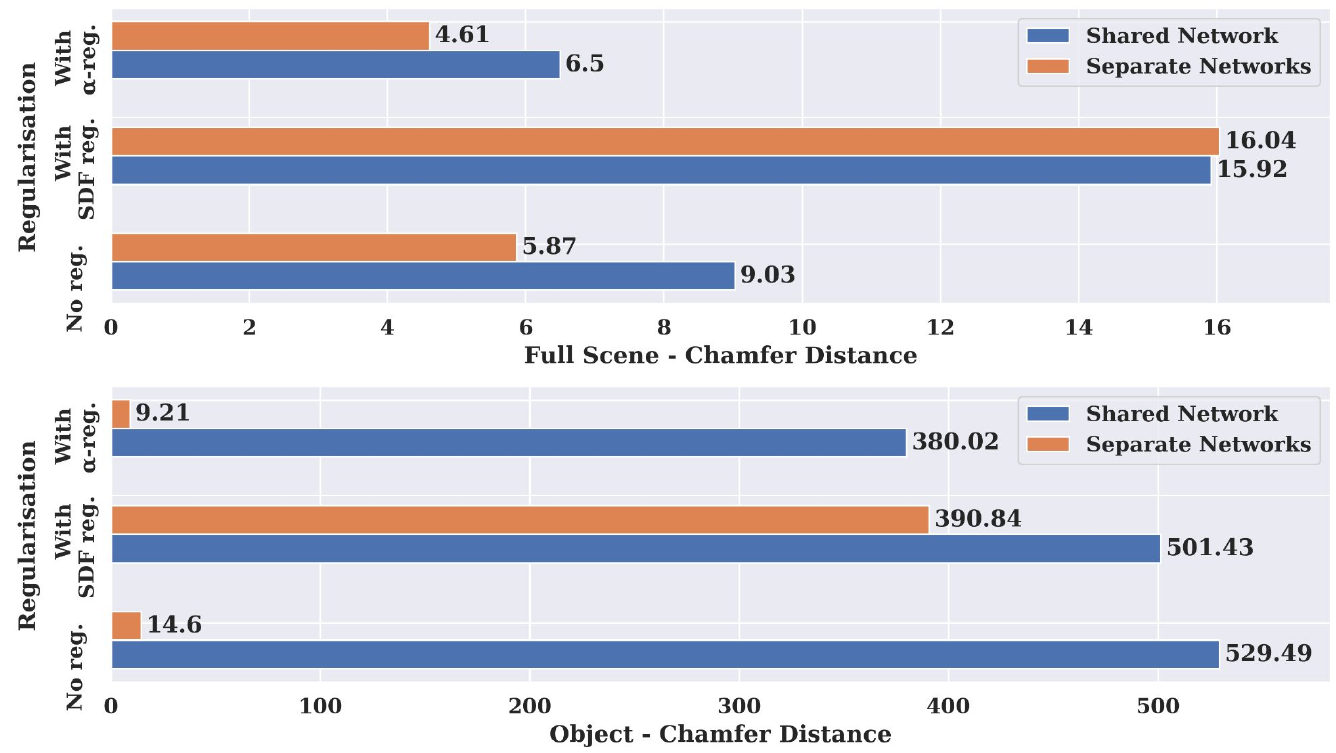}
    \caption[Ablations]{Quantitative evaluation with ablated components shows that having separate MLPs for human and object, and the proposed alpha-regularisation are important for high-quality reconstruction.}
    \label{table:ablations} 
\end{figure}

\begin{figure}
    \centering
    \includegraphics[width=\linewidth]{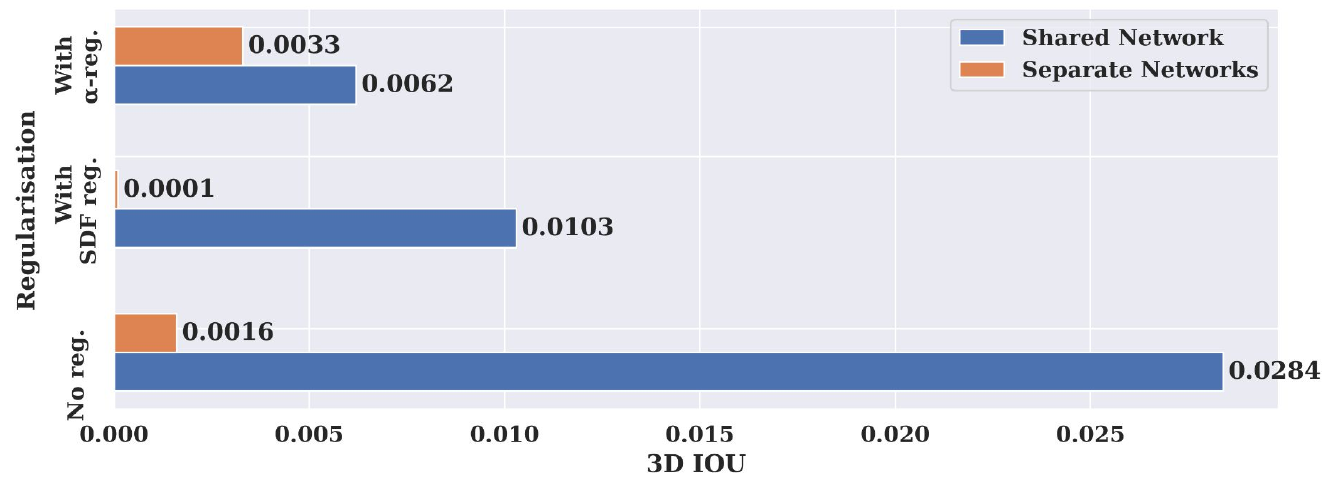}
    \caption[Ablations]{3D IOU to assess the intersection of segmented objects for the ablated components.}
    \label{table:3d_iou_intersection}
\end{figure}

We also perform an ablation study to evaluate the design choices.
In particular, we evaluate the importance of having separate MLPs for the human and the object, instead of a single, shared MLP predicting both the SDFs as done in ObjectSDF++.
We also compare the proposed alpha-regularisation against SDF level regularisation $\sum_p \left(\exp(\frac{\beta}{\lambda_t} \cdot \max(-\Phi_1, 0) \cdot \max(-\Phi_2, 0) \right)$, such that both SDFs are not negative at the same position as mentioned in \cref{sec:losses}. 
The differences in the results are shown in~\cref{fig:ablations}, and the quantitative results are shown in~\cref{table:ablations}.
The overall scene reconstruction largely remains the same, but the individual object and human reconstruction qualities deteriorate because of phantom blobs that get formed underneath the surface of the complimentary SDF when we have shared MLP or we do not use the alpha-regularisation. 
Since these blobs are underneath the surface, they are invisible in renderings, thereby satisfying rendering losses. 
To, demonstrate that the high Chamfer distance is due to the intersecting phantom blobs, we also show results by calculating the 3D IOU between the human and the object. 
Ideally, we do not want any penetrations, hence the IOU should be as low as possible. 
Thus, we can see that having separate MLPs for humans and objects and the proposed alpha-regularisation are important for high-quality reconstruction.

\section{Conclusion}
We introduced a novel method for separable 3D reconstruction of two-object interaction in a multi-view setting considering the challenges of occlusion and difference in object scales. Following the insight that the opaque objects in the scene must have non-overlapping opacities in the implicit network, we showed that the proposed $\alpha$-blending regulariser can indeed incentivise the network to learn disjoint opacities ensuring that the object boundaries remain separate. Through comprehensive experiments, our approach demonstrated the suitability and high accuracy, both on 3D and novel view synthesis metrics and across several datasets. Our simple yet effective regularisation strategy demonstrated in a \textit{generalised} setting can potentially be applied to specific use cases such as template-based human performance capture, or compositional scene generation. We hope that the newly recorded datasets will allow researchers to make further progress in studying the challenging problem of multi-view compositional 3D scene reconstruction. In the future, we intend to refine our method for larger-scale and multi-object scenes and use it for markerless dataset collection. 

{
    \small
    \bibliographystyle{ieeenat_fullname}
    \bibliography{main}
}
\clearpage
\maketitlesupplementary

\begin{figure}    
    \includegraphics[width=\linewidth]{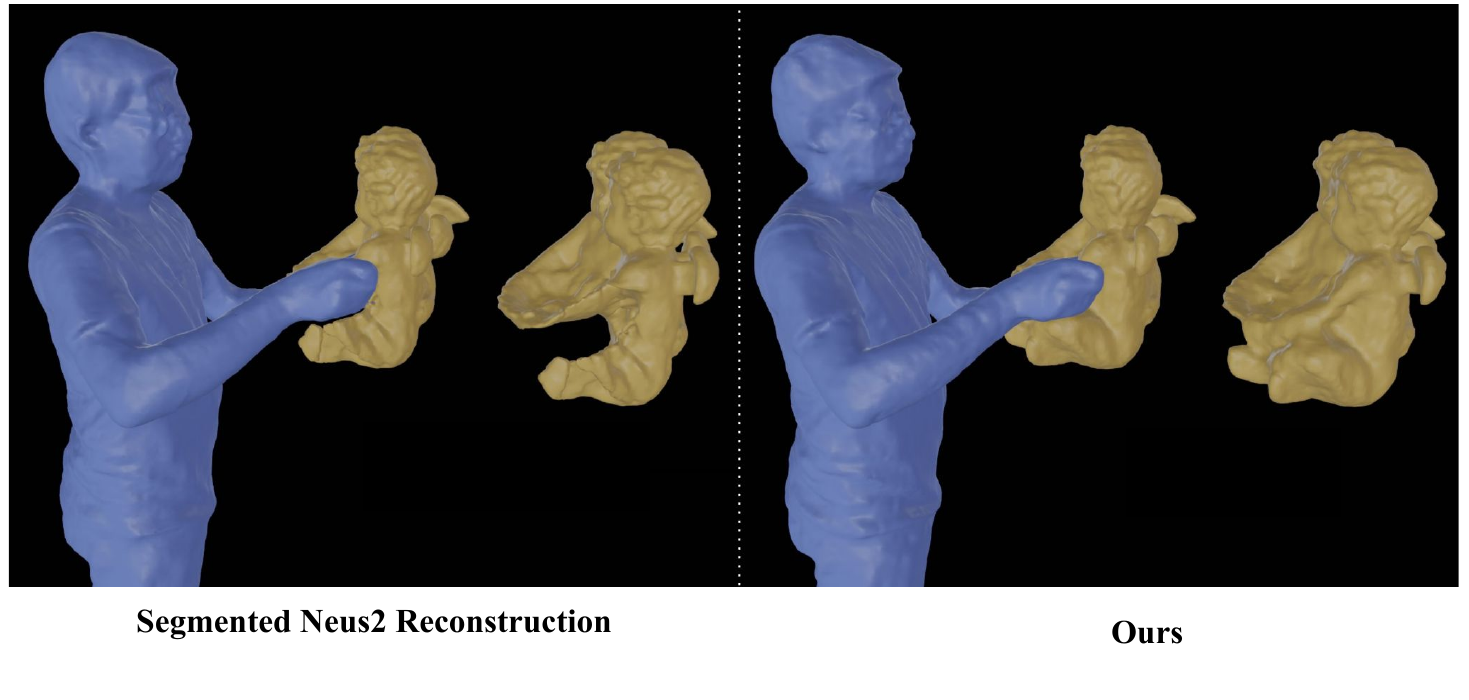}
    \caption{Na\"ively reconstructing the human and the object SDFs using separate NeuS2~\cite{neus2} reconstruction leads to extreme geometric artefacts due to occlusion. 
    }
    \label{fig:naive_sdf}
    \vspace{-1em}
\end{figure}

In this document, we provide more information regarding our implementation in \cref{sec:implementation_details}, more details about our dataset in \cref{sec:human_object_dataset}, a possible extension to the method by incorporating template priors in \cref{sec:object_template_refinement} and discussion about improvements of our method over ObjectSDF++ and ``Segmented Neus2'' in \cref{sec:on_objectsdf++} and \cref{sec:on_segmented_neus2}, respectively. We also present additional qualitative comparisons in \cref{sec:volsdf_comparison} and \cref{sec:obsdf_comparison}.

\section{Implementation Details}
\label{sec:implementation_details}
\noindent \textbf{Scene Encoding:} The sampled point positions are encoded using the hashgrid encoding, $h(\mathbf{x})$, with $L = 18$ levels, two features per level and use a hashmap of size $2^{19}$.
We set the base resolution to 16 and the highest resolution to 8192. Following~\cite{mueller2022instant}, we maintain an occupancy grid of resolution $128$ and skip the empty space while ray marching, whenever the opacity is below $10^{-4}$. The view direction $\mathbf{v}$ is encoded as spherical harmonics up to degree 4. We also use per-image latent of size 8, to account for slight color variations in zoomed-in camera views.
\par
\noindent \textbf{MLPs:} The MLPs for human SDF $\Phi_h$, the object SDF $\Phi_o$ and the feature extractor consist of two layers with 64 neurons each. The colour MLP $\mathcal{C}_s$ is also a 2-layer MLP but with 128 neurons each. 
\par \noindent \textbf{Sampling:} We sample rays for each image in two ways: (1) from pixels within the segmentation masks,  and (2) randomly from any pixel in the image. 
The probability of sampling rays from the masks is progressively increased (as training progresses), from 0.1 to 0.8, linearly increasing from steps 0 to 5000. From steps 0 to 5000, we sample equally from both the human and the object, and after step 5000, we sample randomly from the whole foreground mask.
\par \noindent \textbf{Training:} We train our method for $10k$ steps for each scene, which  takes $\approx$30--45 minutes on a single A40 GPU.
\par \noindent \textbf{Changes to ObjectSDF++:} We increase the total number of hashgrid levels, hashmap size, and resolution to match our implementation, as explained above. ObjectSDF++ also uses depth and normal supervision, since they show their method on indoor scenes. As we do not use either of them, for a fair comparison, we set the normal and depth loss weights to 0. Apart from these, we retain all the other hyperparameters as it is in their implementation. To complete training on one scene, ObjectSDF++ takes around 12--14 hours on a single A40 GPU.

\par \noindent \textbf{Color MLP:} Rather than using a single colour MLP, another option would be to use two separate colour MLPs for each object. But by doing so, each colour network has the freedom to learn the background (or the other object) as colour (black), instead of relying on opacity to give the accumulated colour as 0, in \cref{eq:color_compositing}. Instead, using a single colour MLP ensures that for any position that is occupied by either of the objects, the colour network predicts the correct colour, but overall accumulation depends on the corresponding object opacity being one, and the other opacity being zero.
\par \noindent
\textbf{Segmentation Masks:} We obtain segmentation masks for the human-object, human-human and object-object (WildRGBD) datasets using a pipeline of GroundingDINO~\cite{liu2023grounding} and Segment-Anything~\cite{kirillov2023segany} implemented in ~\cite{langsam}. We further add a CLIP~\cite{CLIP} similarity based filtering, when multiple masks are predicted. Since hand-object (AffordPose~\cite{affordpose}) is synthetic dataset, we get the ground-truth segmentation masks while rendering the meshes.
\section{Human-Object Dataset}
\label{sec:human_object_dataset}
\begin{figure}
    \centering
    \includegraphics[width=\linewidth]{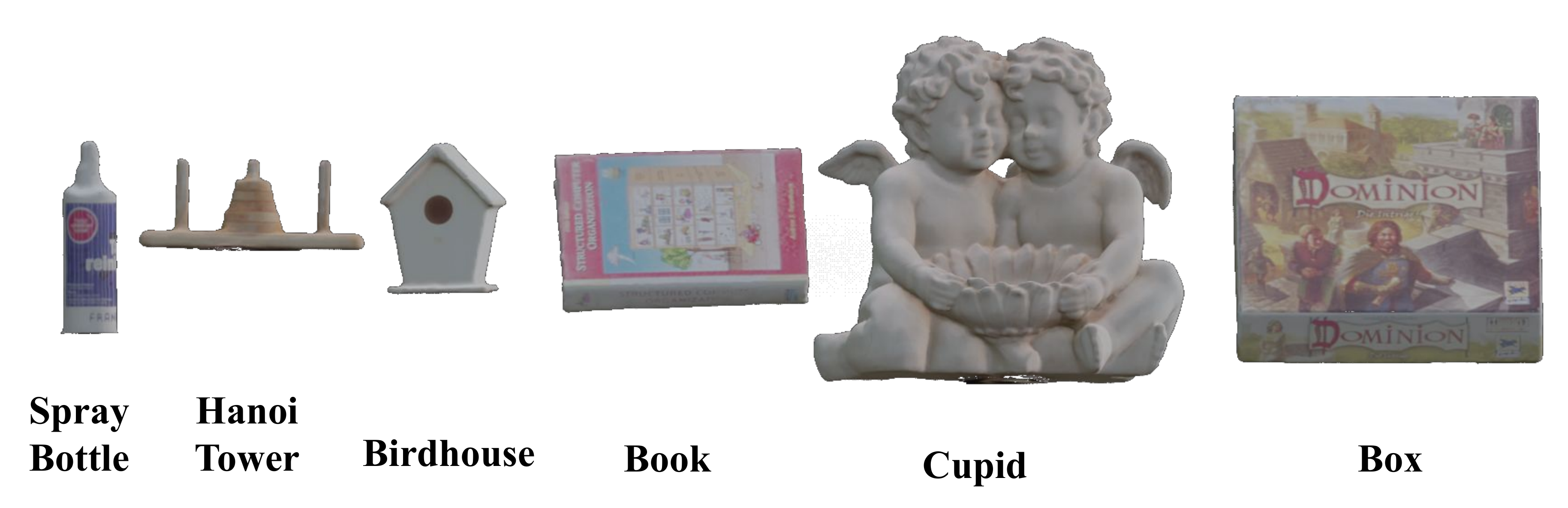}
    \caption{We capture human interactions with six objects of varying intricacy (book vs Hanoi tower) and scale (spray bottle vs sculpture). Yet, the overall scale of the objects remains comparably small. 
    }
    \label{fig:scanned_templates}
\end{figure}

Our new human-object dataset consists of 3 different people each with 6 objects shown in \cref{fig:scanned_templates}. Each scene consists of a maximum of 120 views (some views might be removed because of bad segmentation) with many scenes containing views zoomed into the occupied area. More specifically, for subject 0, all scenes except 'Cupid' have only normal views, and for subject 0 'Cupid' scene, as well as all scenes of subjects 1 and 2, have 19 zoomed-in views. Irrespective of the zoom, all images have been cropped to a resolution of $1200\times1600$ px. 
\section{Limitations}
\begin{figure}[!h]
    \centering
    \includegraphics[width=\linewidth]{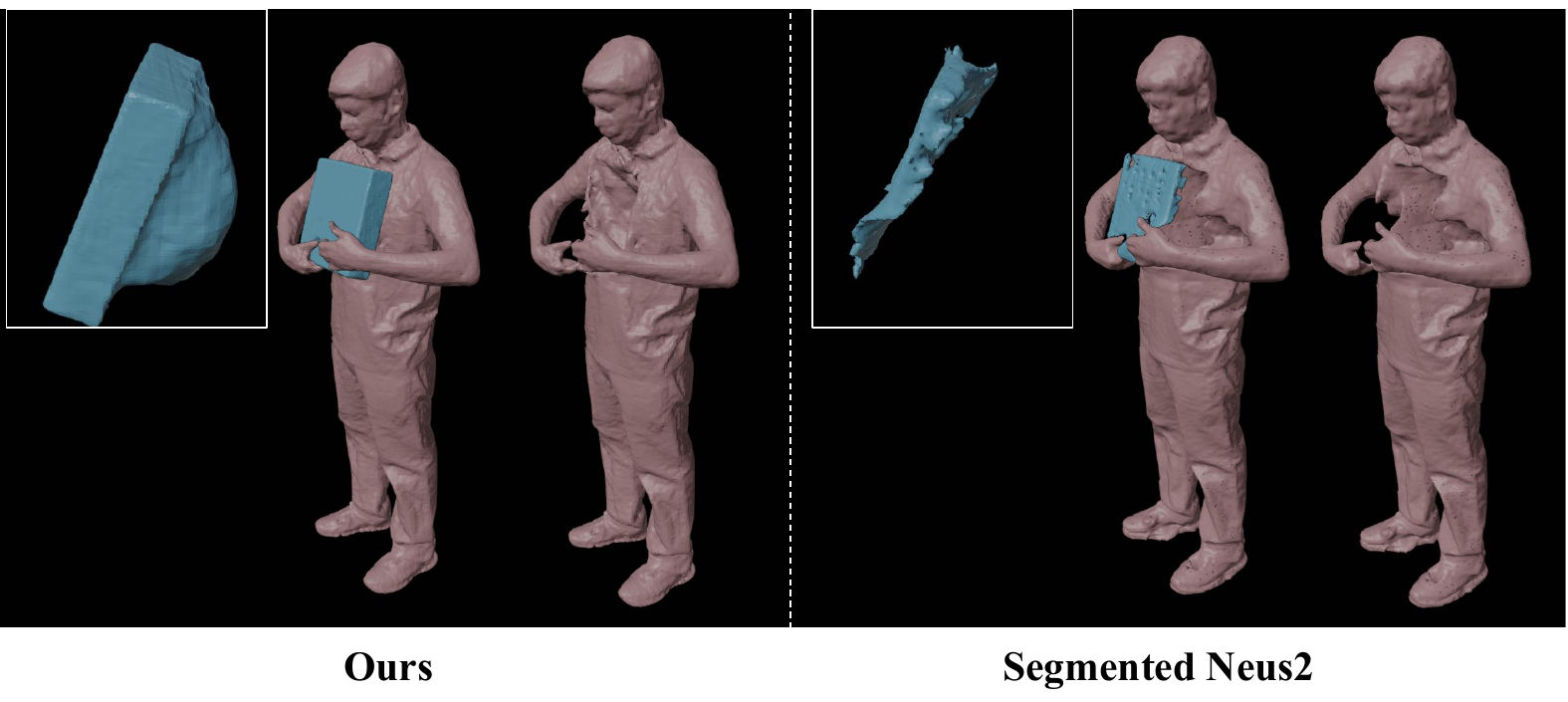}
    \caption{Failure case: under heavy occlusion, our method generates extended, yet separate geometries. Segmented NeuS2, on the other hand, reconstructs the scene with holes.}
    \label{fig:failure_case}
\end{figure}

Our method is designed to separately reconstruct two interacting objects. 
While this is largely addressed by encoding both geometries in a shared hash grid and ensuring the opacities are disjoint, some artefacts remain.
Consider the case in~\cref{fig:failure_case}: 
The object's face towards the body is occluded beyond observation.  
Hence, the method does not have sufficient prior to disambiguate the human-object boundaries.  
Yet, it ensures that the boundaries are separate. 
A potential solution to this problem would be incremental training, as shown in~\cref{fig:incr_failure_case}, or fine-tuning the joint SDF with a pre-scanned template providing a useful geometric prior, assuming it is available. 
Similar refinement can be done for the human; see the discussion in \cref{sec:object_template_refinement}
Another limitation is in cases of thin gaps between different structures, as in the hand fingers of the \textit{Bag}            scene shown in~\cref{fig:hand_object_geometry}. 
Sometimes, we observe undesired \textit{bridges} between such thin gaps due to the nature of ray sampling during optimisation. 

\section{Consecutive Frame-by-Frame Reconstruction} 
\label{sec:incr_training}
\begin{figure}[h]
    \centering
    \includegraphics[width=\linewidth]{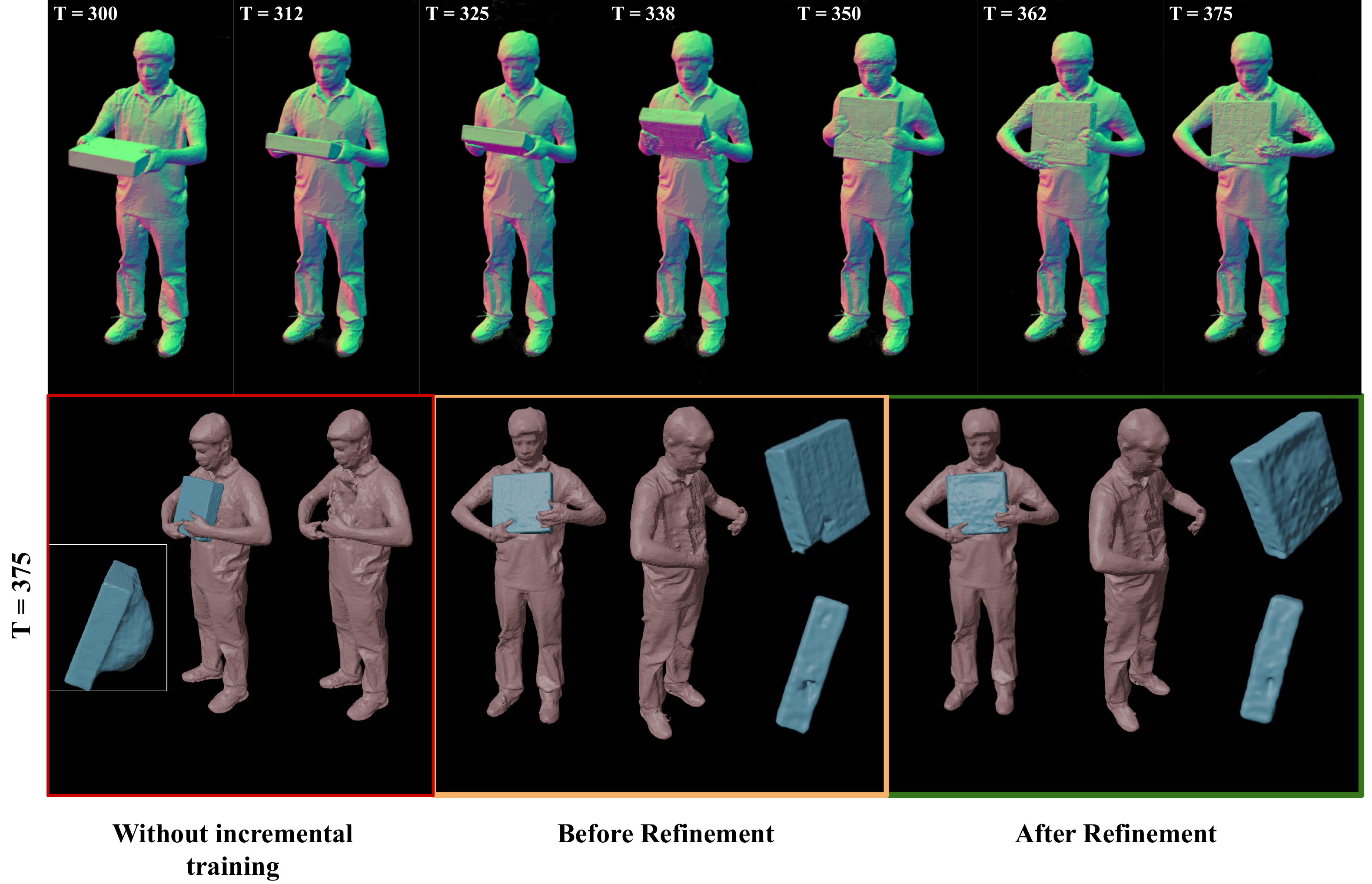}
    \caption{
    Improvements in the 3D object geometry using incremental training. 
    We observe that the largest erroneous object deformations caused by heavy occlusions are mostly corrected using incremental training.}
    \label{fig:incr_failure_case}
\end{figure}

Our approach can also be applied on consecutive frames by initialising the parameters for the current frame from the previous frame. 
This also helps prevent certain defects, as the model has prior on the shapes observed before occlusions. 
One such example is presented in \cref{fig:incr_failure_case}, which improves the failure case \cref{fig:failure_case}. \cref{fig:incr_failure_case} shows normal renderings for a few sampled time steps.
\label{sec:suppl_incr_training}
\section{Additional Discussion}
\subsection{On Object Templates}
\label{sec:object_template_refinement}
\begin{figure}
    \centering
    \includegraphics[width=\linewidth]{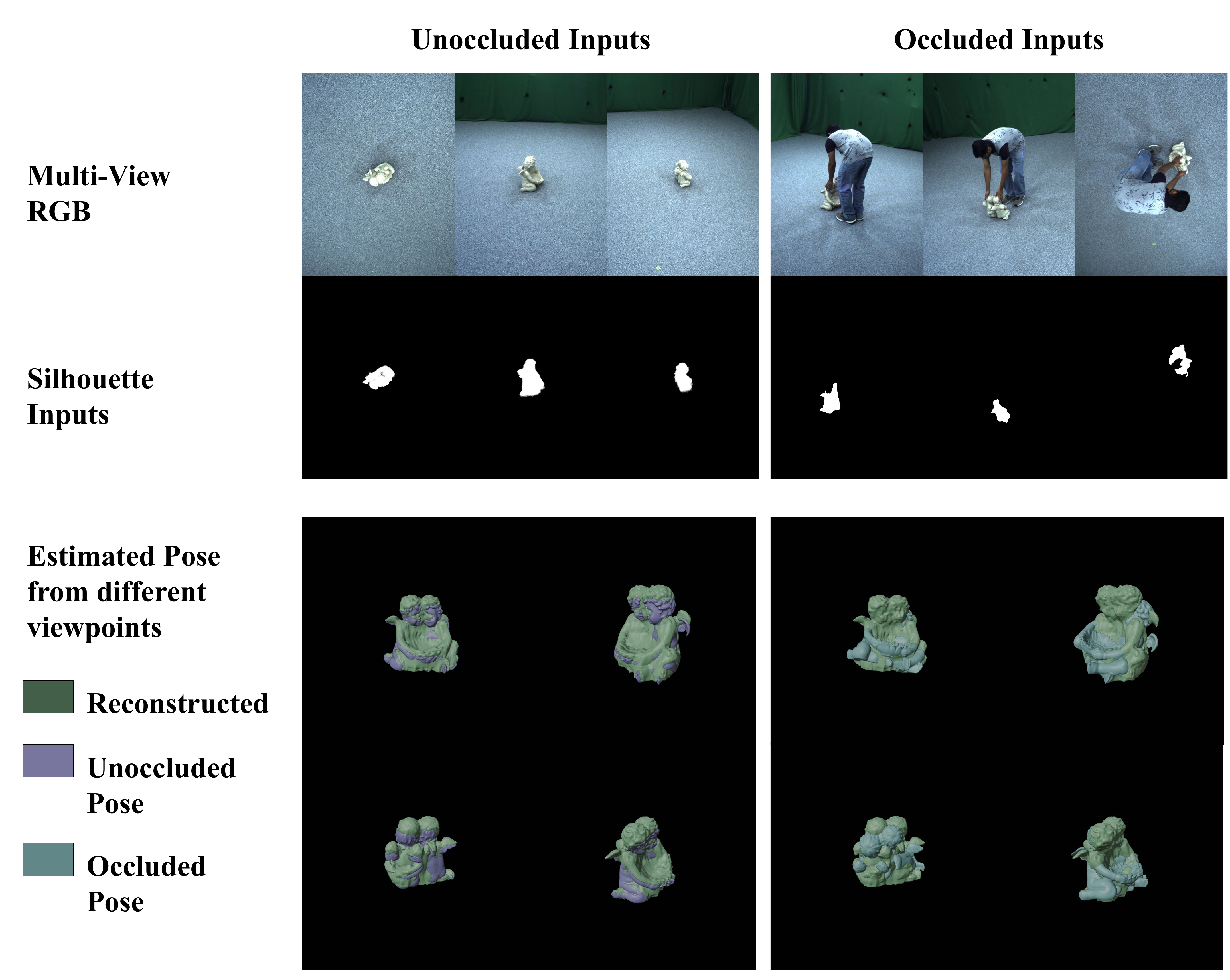}
    \caption{Comparison of occluded and unoccluded 6DOF fitting a template using Silhouette loss (initialised with known position). The orientation of the fitted template is compared against the reconstructed mesh.}
    \vspace{-1em}
    \label{fig:pose_estimation}
\end{figure}
In this work, we assume that an object template is not available. 
While having a template, arguably, would make the task easier in an alternative setting, it would also substantially limit the method's applicability and extendability to scenarios with arbitrary objects. 
It would also necessitate the additional step of template acquisition, which can be infeasible in many downstream applications. 
Moreover, articulated objects like laptops would require a different approach to 3D reconstruction, even when the template in a canonical pose is available; similar observations apply to humans. 
Hence, we focus on modelling two-object interactions at a fundamental level which can, if required, be extended if the template is available. 
We next briefly discuss several considerations in this regard. 
\par
Suppose an object template is available. 
How could our approach be extended or adjusted to account for this prior knowledge? 
A naive way would be to fit the object template to the image observations using 6DoF optimisation. 
This is, however, suboptimal since  
(1) the colour rendering loss cannot be used because the lighting conditions at the time of template acquisition and 6DoF optimisation would be likely different; 
(2) for the same reason as mentioned above (i.e.~since the template appearance is likely to differ during the template acquisition and the main scene capture steps), the globally optimal 6DoF template pose could be inaccurate; and (3) the silhouette-based optimisation would also struggle as the objects are under severe occlusion, and the segmented silhouettes are not reliable as shown in Fig~\ref{fig:pose_estimation}.
\par
A better alternative would be to fit the template pose coarsely to the scene and use the posed template to sample the points for volume rendering. 
This is akin to the \textit{canonicalised} representation in several 3D human and non-rigid reconstruction works~\cite{pumarola2020d, liu2021neuralactor, tretschk2021nonrigid, peng2021neuralbody}. 
While this would improve the convergence speed, such an approach would still benefit from the shared representation and alpha-blending loss proposed in this work. 
\par
\begin{figure}
    \centering  
    \includegraphics[width=\linewidth]{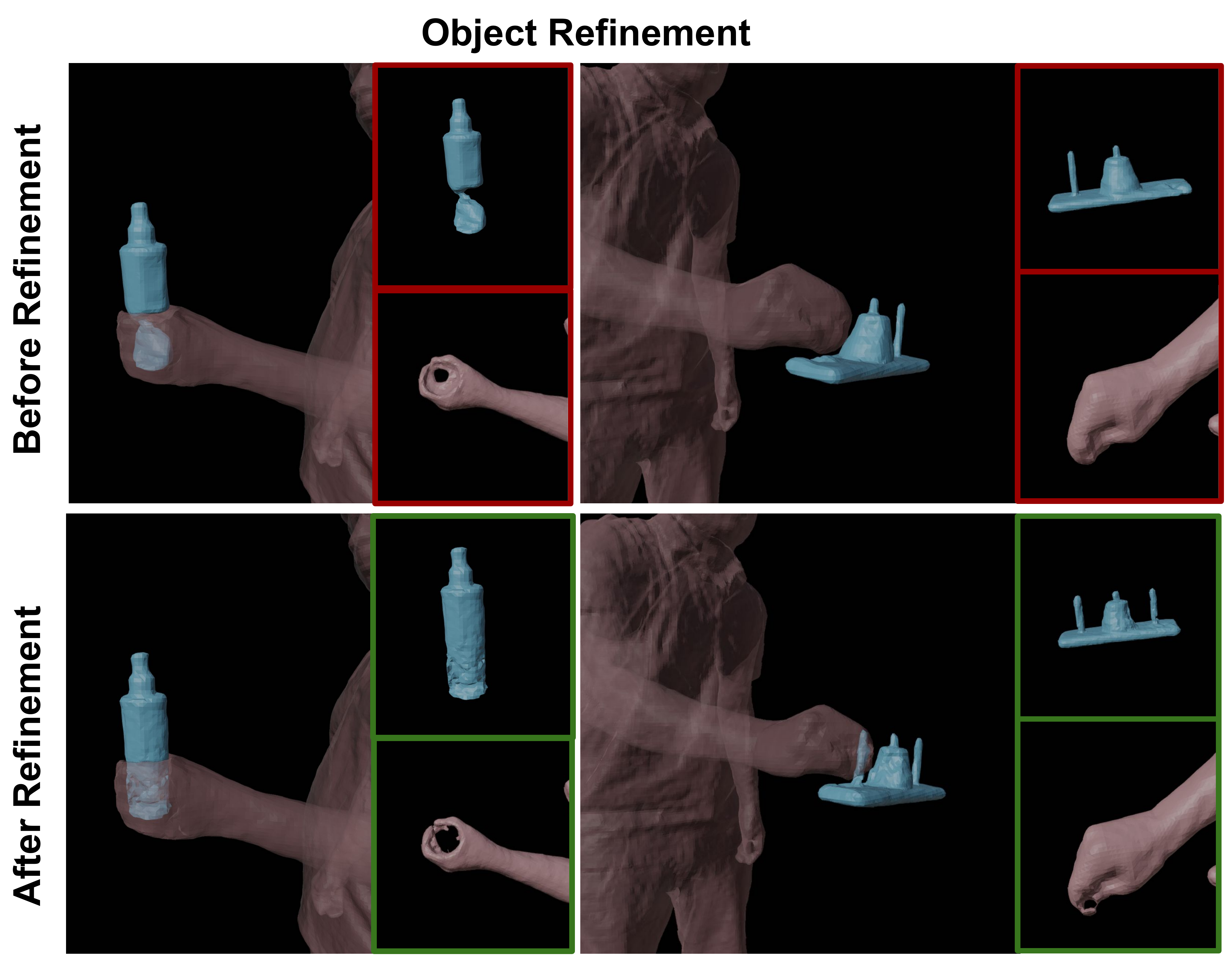}
    \label{fig:object_refinement}
    \caption{Illustrations of the object's geometry before and after the template-guided refinement. Notice that the peg of the tower, missing in the first state, re-emerges after jointly optimizing with the template. Interestingly, jointly optimizing with the template also improves hand reconstruction, as can be seen in the case of spray holding.}
    \vspace{-1em}
\end{figure}

Another potential alternative would be instead to fit the object template using the object's reconstructed surface SDF. 
The optimisation would be performed in two alternating steps until convergence, i.e.~(step 1) using the object SDF to update the template's pose and (step 2) using the optimised template pose to refine the joint scene geometry with the help of our method to alleviate penetration artefacts. 
Indeed, we observe that this joint optimisation improves scene reconstruction, especially in the case of human-object interaction. 
As shown in \cref{fig:object_refinement}, the hand geometry benefits from template-guided optimisation using this iterative refinement policy.
\par
\subsection{On ObjectSDF++}
\label{sec:on_objectsdf++}
ObjectSDF++ is the closest work to our proposed method. 
There are, however, two key differences that allow us to outperform ObjectSDF++ across multiple evaluation settings.
First, ObjectSDF++ uses a single(shared) MLP for all SDF outputs, whereas we model each SDF with a separate MLP. 
This introduces a tradeoff -- better reconstruction and separation quality at the cost of the ability to model multiple-objects. 
This is also confirmed by the ablations presented in the main draft. 
It is noteworthy that our solution can also be extended to multiple-objects by having multiple MLPs, in-theory. 
This would require alpha-regularisation on all combinations and is a direction for future exploration.
Second, ObjectSDF++ proposes a ReLU-based regularizer which, we hypothesize, is harder to optimize. 
This is evident in Fig.~5 and Fig.~7 (main) wherein some ObjectSDF++ reconstructions have deeper deformations near contact regions than ours.
Finally, ObjectSDF++ is additionally trained using depth and normal maps whereas we do not need such supervision.

\subsection{Reasons for Better Reconstruction Compared to Segmented Neus2}
\label{sec:on_segmented_neus2}
NeuS(2) is sensitive to occlusions in the input. Occlusion in one view implies all rays along the entire path hit blank space, whereas other views indicate the same space is non-empty. This mismatch leads to incorrect optimisation in the form of artefacts and holes. On the other hand, by jointly encoding and rendering both geometries through a shared hashgrid, we can maintain multi-view consistency, since the presence of one object explains the absence of the other object from a particular viewpoint. This is further improved by introducing alpha-regularisation, which prevents penetrations.
\section{Experiments (Continued)}
\subsection{Hand-Object}
\begin{figure*}
    \centering
    \includegraphics[width=\linewidth]{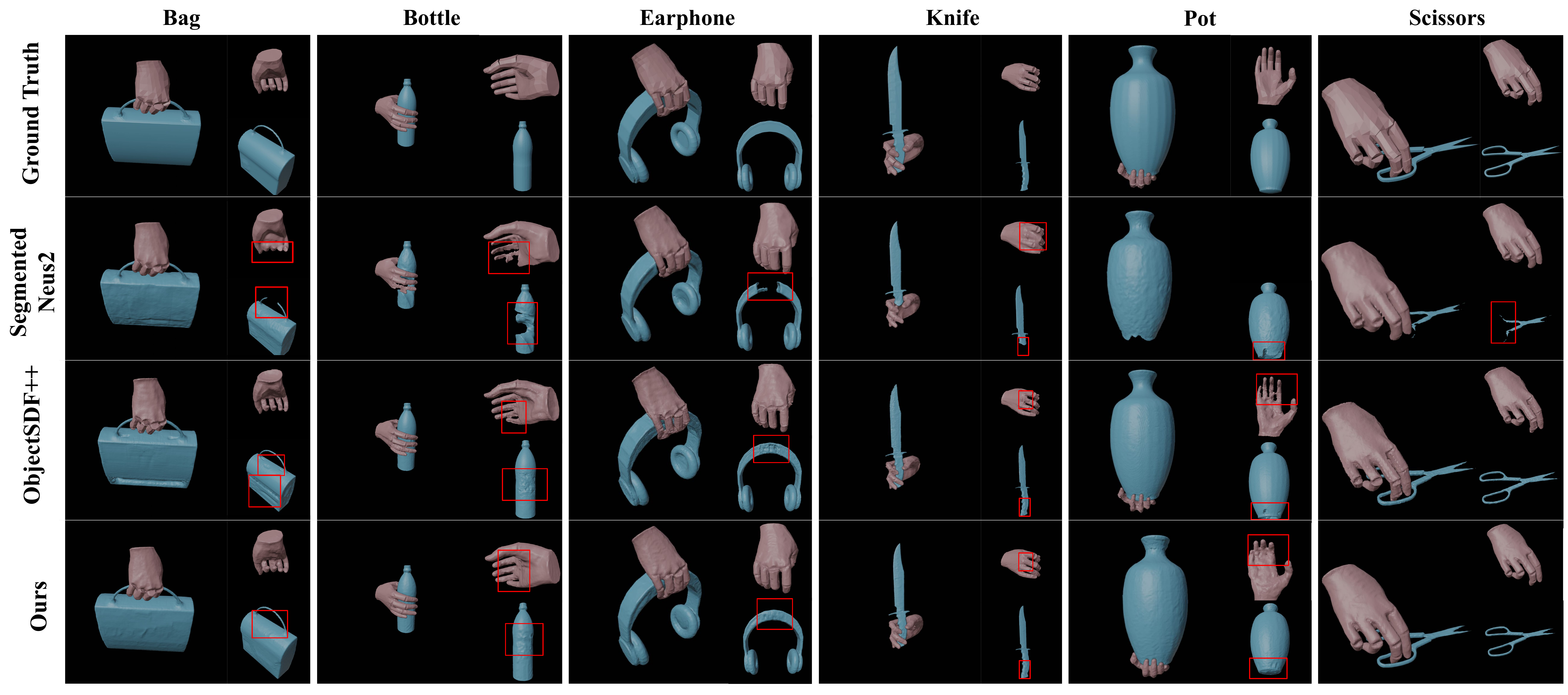}
    \caption{Qualitative comparison on the  AffordPose dataset. The regions highlighted in red indicate apparent differences in the reconstructions. In the case of Segmented Neus2, for the pot scene, reconstruction of the hand fails (hence not shown). \textbf{Best viewed when zoomed.}
    } 
    \label{fig:hand_object_geometry}
    \vspace{-1em}
\end{figure*}
We show the qualitative comparison for hand-object scene geometry reconstruction in \cref{fig:hand_object_geometry}.
\subsection{Human-Human Appearance Evaluation}
\label{sec:human_human_appearance_suppl}
\begin{figure*}[h]
    \vspace{-20pt}
    \includegraphics[width=0.95\linewidth]{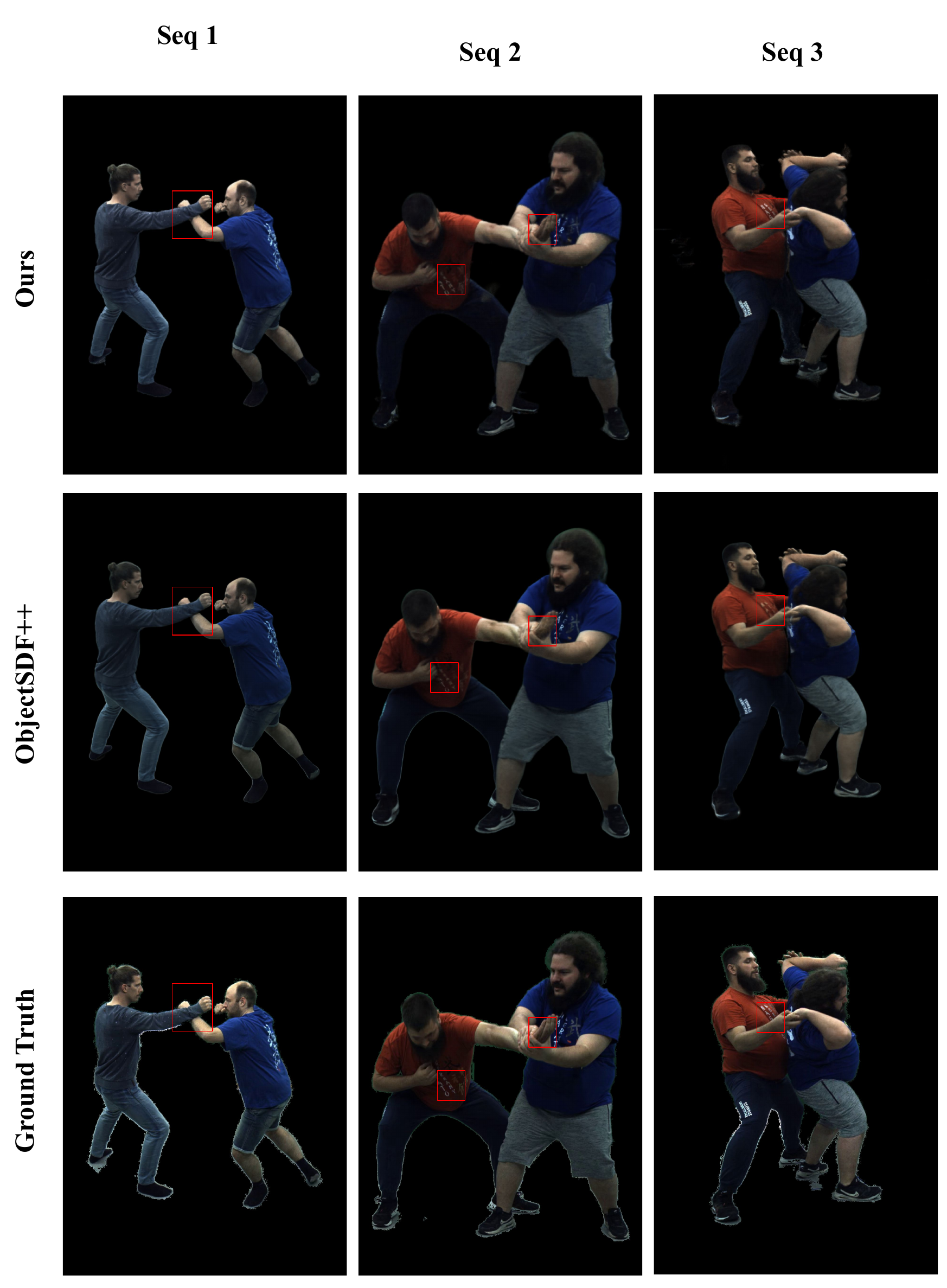}
    \caption{Qualitative comparison with ObjectSDF++ of human-human interaction novel-view synthesis.}
    \label{fig:human_human_appearance}
\end{figure*}

We also show qualitative comparison for novel-view synthesis on human-human interaction scenes in \cref{fig:human_human_appearance} and quantitative comparison in \cref{table:human_human_appearance}.

\begin{table}
\resizebox{\columnwidth}{!}{
\begin{tabular}{|c|c|c|c|c|c|c|}
\hline
\textbf{Seq ID} & \multicolumn{2}{c|}{\textbf{PSNR$\uparrow$}}    & \multicolumn{2}{c|}{\textbf{SSIM$\uparrow$}}        & \multicolumn{2}{c|}{\textbf{LPIPS$\downarrow$}}       \\ \hline
\textbf{}         & \shortstack{\textbf{Object} \\ \textbf{SDF++}} & \textbf{Ours} & \shortstack{\textbf{Object} \\ \textbf{SDF++}} & \textbf{Ours} & \shortstack{\textbf{Object} \\ \textbf{SDF++}} & \textbf{Ours} \\ \hline
1     &   25.84                   & \gold{30.50}                          &  0.92                                         & \gold{0.94}                           &  0.14                                         & \gold{0.13}                            \\ \hline
2    & 29.12              & \gold{31.90}                         & \gold{0.95}                                 & \gold{0.95}                           & \gold{0.12}                                    & 0.13                           \\ \hline
3    & 24.20           & \gold{28.66}                          & 0.91                                 & \gold{0.92}                           & 0.18                                 & \gold{0.17}                           \\ \hline
\end{tabular}}
\label{table:human_human_appearance}
\vspace{-1em}
\caption{Quantitative comparison of our method with the ObjectSDF++ on the novel-view synthesis task of the Human-Human scenes.}
\end{table}

\subsection{NeuralDome Results}
We show a qualitative comparison on a human-table interaction scene from NeuralDome \cite{zhang2023neuraldome} dataset. While NeuralDome uses pre-scanned template of the object, along with markers, our method can obtain a similar reconstruction quality using only multi-view images.
\begin{figure}
    \centering
    \includegraphics[width=\linewidth]{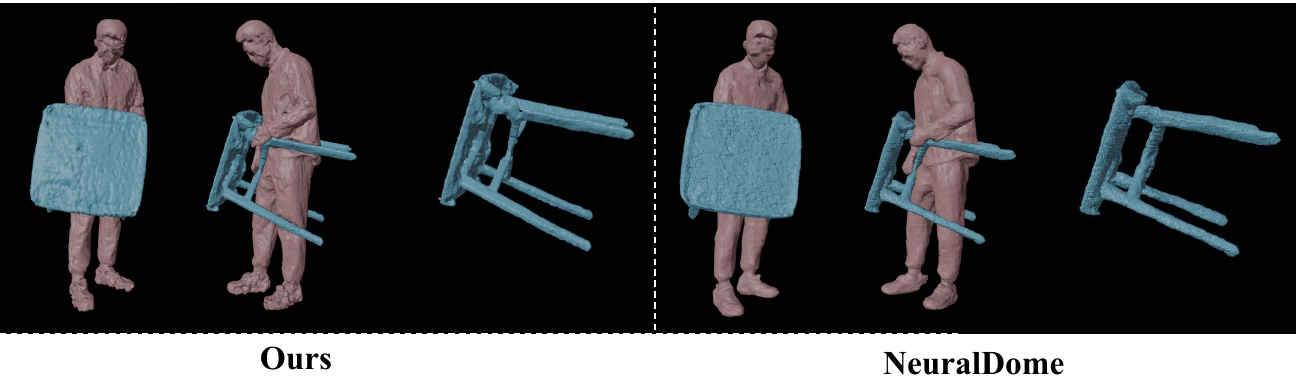}
    \caption{Reconstruction comparison with a scene from the NeuralDome dataset. NeuralDome provides a pre-scanned template of the object and a multi-view 3D reconstructed human.} 
    \label{fig:neural_dome_table}
\end{figure}

\section{Comparison against VolSDF}
\label{sec:volsdf_comparison}
\begin{figure}[!h]
    \centering
    \includegraphics[width=\linewidth]{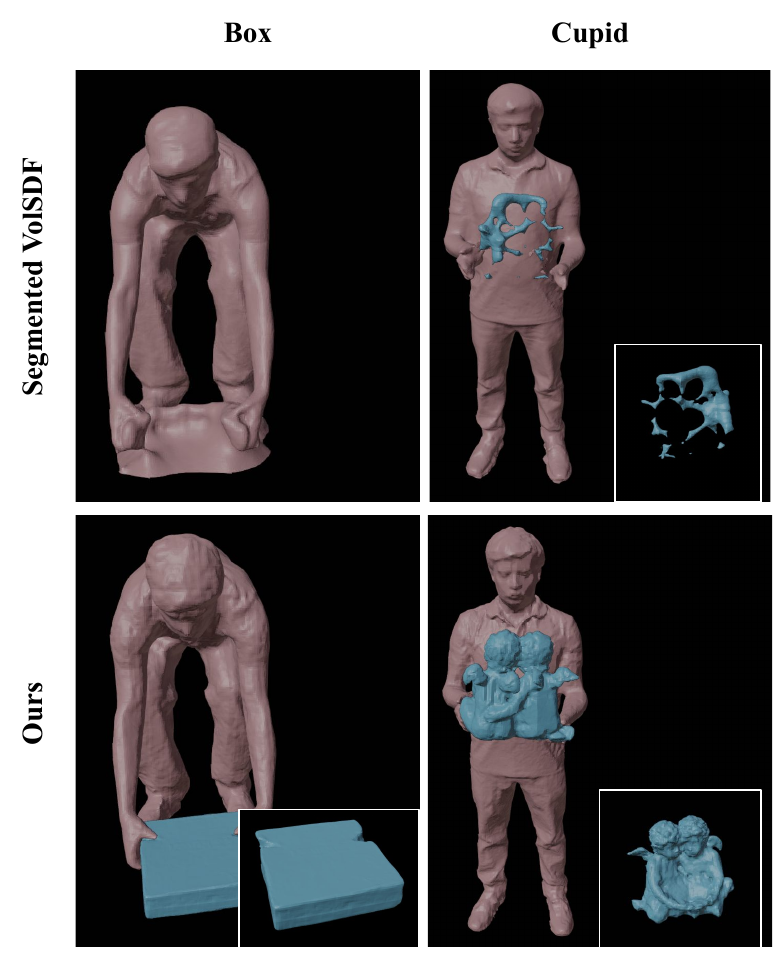}
    \caption{Qualitative comparison with Segmented VolSDF. We observe that similar to the Segmented NeuS2, the object is not reconstructed reasonably.} 
    \vspace{-1em}
    \label{fig:volsdf_comparison}
\end{figure}
Recall that we show results with ``Segmented NeuS2" by training two different NeuS2 models for the human and object.
This makes the geometric reconstruction of the object agnostic of the presence of a human and vice-versa.
In order to confirm that the failure of reconstruction for the "Segmented Neus2" is not just because of NeuS \cite{wang2021neus} formulation, we also use VolSDF \cite{yariv2021volume}, and train it in isolation for human and object, by providing the respective masks. 
We show the results on the two biggest objects in our evaluation dataset, i.e.,~Box and Cupid statue, in \cref{fig:volsdf_comparison}. While human reconstruction works rather well, the box is not reconstructed and the cupid is reconstructed poorly. 
This demonstrates, yet again, that the baseline approach of two isolated reconstructions is suboptimal and that sharing the scene parameters is crucial to separable reconstruction. 

\section{Comparison against ObjectSDF}
\label{sec:obsdf_comparison}
\begin{figure}
    \includegraphics[width=\linewidth]{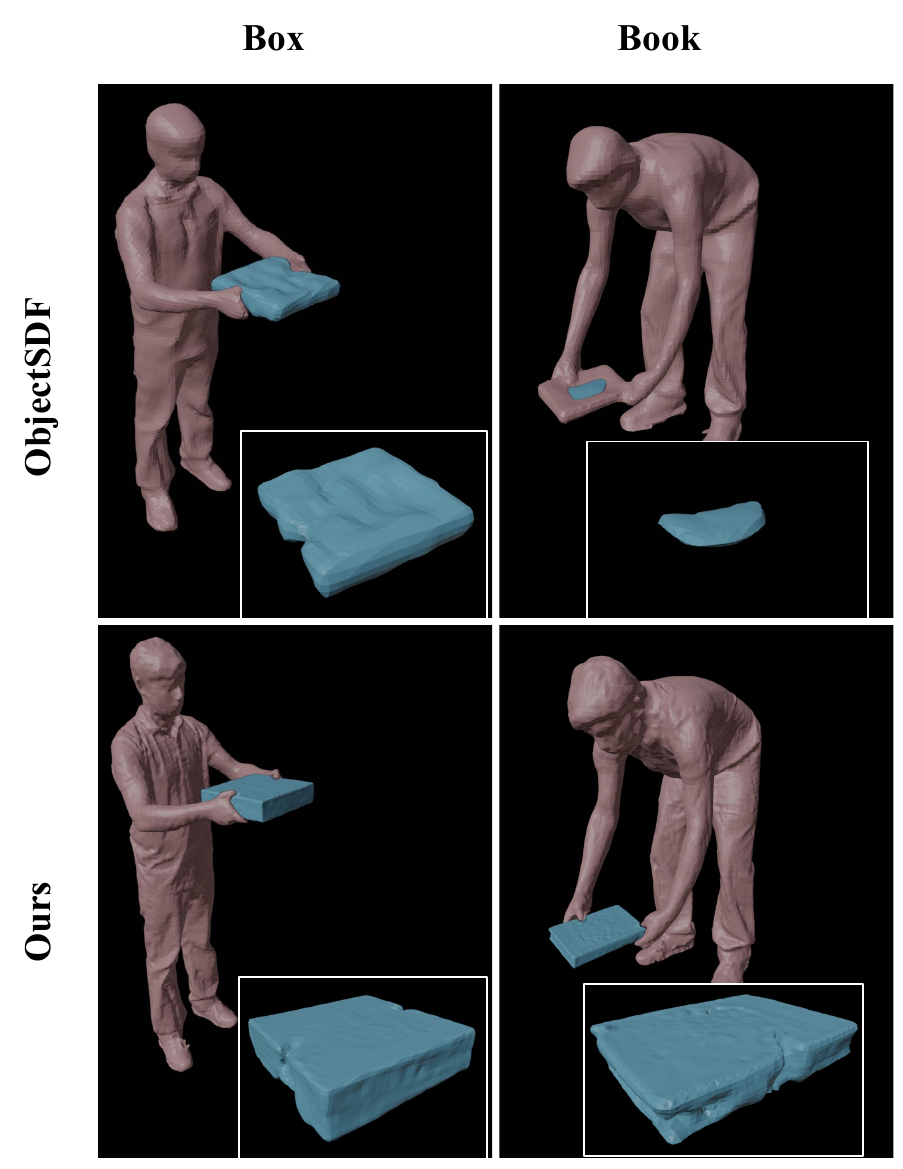}
    \caption{Qualitative comparison with ObjectSDF. We observe that our reconstruction results are much more detailed and well separated, whereas ObjectSDF produces incorrect geometry for the object, especially for the book. 
    } 
    \vspace{-1em}
    \label{fig:objectsdf_comparison}
\end{figure}

We also compare against ObjectSDF \cite{wu2022object} (which was the predecessor to ObjectSDF++) for a few scenes and show the qualitative results in \cref{fig:objectsdf_comparison}. 
Note that ObjectSDF inaccurately assigns large parts of the book to the human (in red). 
The actual book (in blue) is poorly reconstructed. 

\end{document}